\documentclass{article}

\PassOptionsToPackage{numbers, sort&compress}{natbib}

\usepackage[eandd, final]{neurips_2026}

\usepackage[utf8]{inputenc} 
\usepackage[T1]{fontenc}    
\usepackage[
    colorlinks=true,
    linkcolor=blue!60!black,
    citecolor=blue!60!black,
    urlcolor=blue!60!black
]{hyperref}      
\usepackage{url}            
\usepackage{booktabs}       
\usepackage{amsfonts}       
\usepackage{nicefrac}       
\usepackage{microtype}  
\usepackage{xcolor}
\usepackage{fontawesome}
\usepackage{xspace}
\usepackage{hwemoji}
\usepackage{amsmath}
\usepackage{wrapfig}  
\usepackage{colortbl}
\usepackage{cleveref}
\usepackage{multirow}
\usepackage{subcaption}
\usepackage{tcolorbox}
\usepackage{listings}
\tcbuselibrary{breakable, skins, listings}

\tcbset{
    promptbox/.style={
        breakable,
        enhanced,
        colback=gray!8,
        colframe=gray!40,
        fonttitle=\bfseries\small,
        title={Prompt},
        coltitle=black,
        attach boxed title to top left={yshift=-2mm, xshift=4mm},
        boxed title style={colback=gray!20, colframe=gray!40},
        left=6pt, right=6pt, top=6pt, bottom=6pt,
        fontupper=\small\ttfamily,
    }
}

\newcommand{\ours}{\textsc{VistaQA}\xspace}
\newcommand{\metric}{\textsc{Grove}\xspace}

\newcommand{\eg}{\textit{e.g.}\xspace}

\newcommand{\wrt}{\textit{w.r.t.}\xspace}

\title{\ours: Benchmarking Joint Visual Question Answering and Pixel-Level Evidence

}

\author{
    Mozhgan Nasr Azadani\textsuperscript{\rm 1,2}, 
    Yimu Wang\textsuperscript{\rm 1}, 
    Yongpeng Zhu\textsuperscript{\rm 1,*} , 
    Lihong Chen\textsuperscript{\rm 1,*},
    Milan Ganai\textsuperscript{\rm 2}, \\
    \textbf{Sean Sedwards}\textsuperscript{1}, 
    \textbf{Marco Pavone}\textsuperscript{\rm 2,3,$\dagger$},
    \textbf{Krzysztof Czarnecki}\textsuperscript{\rm 1,$\dagger$}  \\
    \textsuperscript{\rm 1} University of Waterloo, \textsuperscript{\rm 2} Stanford University,
    \textsuperscript{\rm 3} NVIDIA 
}

\begin{document}
\maketitle
\begingroup
\renewcommand\thefootnote{}
\footnotetext{
\textsuperscript{*} Equal contribution. 
\textsuperscript{$\dagger$} Equal advising.
}
\renewcommand\thefootnote{\textsuperscript{}}
\footnotetext{Correspondence to: \texttt{mnasraza@uwaterloo.ca}}
\endgroup

\begin{abstract}
Establishing a clear link between model predictions and the visual evidence that supports them is critical for transparency and reliability in multimodal reasoning, yet current multimodal large language model (MLLM) evaluations do not explicitly enforce this alignment. Existing benchmarks assess either textual answer correctness or pixel-level localization in isolation, leaving the coupling of reasoning and grounding an open challenge. We introduce \ours, a comprehensive benchmark for joint evaluation of free-form answer correctness and pixel-level evidence grounding in visual question answering. \ours comprises 1,157 expert-curated samples spanning six task types and six visual domains, ranging from direct perception to compositional and relational reasoning. \ours requires models to not only answer correctly, but to also provide precise segmentation masks that support their answers. It also includes hallucination-aware examples where no valid visual evidence exists.
To support this enhanced evaluation, we introduce \metric, a unified evaluation metric that enforces joint correctness by combining textual accuracy and grounding quality via a per-sample geometric mean, ensuring neither dimension can compensate for deficiencies in the other.
Comprehensive experiments across grounding-aware models and hybrid pipelines with general-purpose MLLMs reveal that even the strongest systems achieve limited performance under \metric, highlighting a substantial gap between answer accuracy and visual evidence alignment. Project page can be found at \url{https://vistaqa.github.io}.

\end{abstract}

\begin{figure*}[t]
    \centering
    \includegraphics[width=\textwidth]{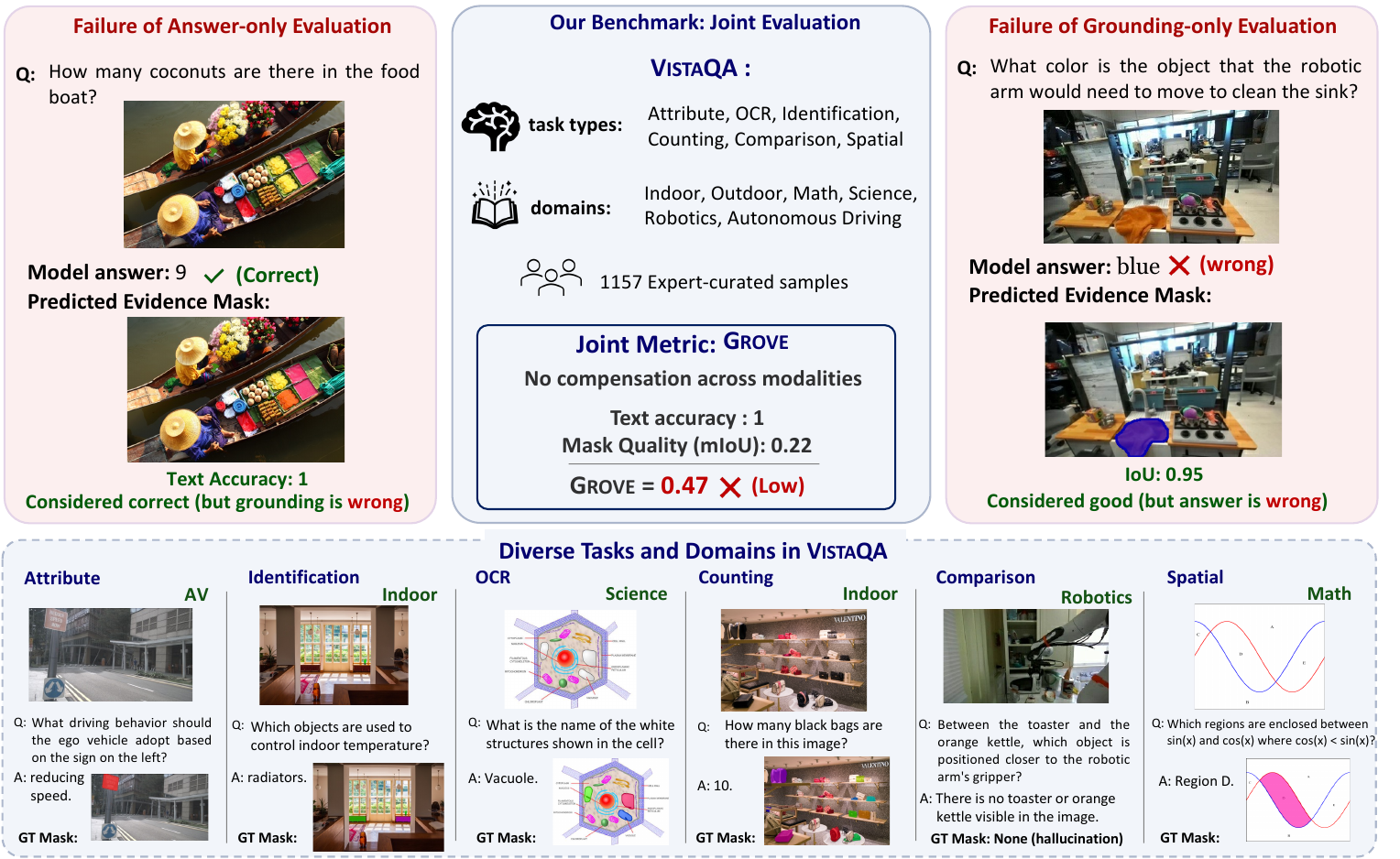}
    \vspace{-1.5em}
    \caption{
        \ours jointly evaluates answer correctness and pixel-level evidence across six tasks and six domains, requiring both to be correct and preventing compensation between modalities.
    }
    \label{fig:teaser}
   \vspace{-2em}
\end{figure*}

\section{Introduction}

Multimodal large language models (MLLMs) have demonstrated remarkable progress across a wide range of multimodal tasks, including visual question answering (VQA), image captioning, and compositional reasoning~\cite{yang2025qwen3, wang2025internvl3, wang2025hawaii, chen2025eagle, chen2024internvl}. Despite these advances, standard evaluation protocols~\cite{liu2024mmbench, yue2024mmmu, qian2024nuscenes} continue to assess performance primarily through textual correctness. A model is rewarded for producing the right answer, even when that answer is not supported by the appropriate visual evidence. Conversely, a model may produce plausible but incorrect answers driven by language priors, despite contradictory visual cues (e.g., answering “two” legs for an animal that visibly has three~\cite{vo2026vision})~\cite{guan2024hallusionbench, liu2025phd, datta2025evaluatingpope}. This creates a fundamental limitation: textual correctness alone cannot determine whether a reply is truly grounded in the image, an essential requirement for model interpretability and hallucination mitigation. For reliable deployment in real-world settings, models must be evaluated on not only \emph{what they say}, but also \emph{what visual evidence} supports their statements.

Recent efforts toward grounding have emerged from two complementary but largely disconnected directions. On one hand, reasoning segmentation models~\cite{lai2024lisa, rasheed2024glamm, wei2024lasagna, wu2024sesame,yan2024visa} demonstrate strong pixel-level localization capabilities, but they are not optimized for generating rich, free-form textual answers and have been reported to achieve limited success on general VQA benchmarks~\cite{zhang2024omg}. On the other hand, state-of-the-art MLLMs~\cite{google2026gemini31, openai2026gpt54} achieve high accuracy on complex multimodal reasoning tasks, but typically do not produce explicit visual evidence to substantiate their predictions. While more recent visual grounding approaches~\cite{yuan2025vrt, wang2026treebench, yuan2025sa2va, qian2025uground} begin to bridge this gap, their textual and visual capabilities are still largely evaluated in isolation: models are scored on answer accuracy and segmentation quality as separate metrics, usually on different datasets (\eg, VQA benchmarks for text and segmentation datasets for masks). Consequently, existing evaluation protocols do not enforce alignment between answers and their supporting visual evidence, leaving the coupling of reasoning and grounding an open challenge.

Existing benchmarks reflect and reinforce this divide. VQA benchmarks~\cite{fu2025mme, yue2025mmmupro, lumathvista, goyal2017vqa2} evaluate compositional reasoning through textual answers, but provide no mechanism to verify whether those answers are grounded in the correct image regions. Conversely, referring expression and reasoning segmentation benchmarks~\cite{lai2024lisa, yu2016modelingrefcoco, sahoo2026converseg} emphasize localization, but treat segmentation as the prediction target rather than as evidence supporting a textual answer. More recent efforts attempt to combine answering and grounding signals; however, grounding is typically treated as a referring or auxiliary output rather than explicit evidence~\cite{wu2024vstar, chen2022vizwizgrounding}, and evaluation protocols assess answer correctness and localization quality independently, without enforcing consistency between them~\cite{yuan2025vrt, wang2026treebench}. In addition, these benchmarks are often limited to narrow domains or task settings and rarely account for hallucination, where models must recognize the absence of valid visual evidence. As a result, there remains no benchmark that systematically requires models to produce correct free-form answers that are explicitly supported by pixel-level visual evidence across diverse tasks and domains.

To address this gap, we introduce \ours, a benchmark for the joint evaluation of free-form answer correctness and pixel-level evidence grounding in VQA. Each sample in \ours consists of a question, a reference answer, and a segmentation mask that specifies the visual evidence required to support that answer. Current state-of-the-art MLLMs are insufficiently reliable to autonomously generate complex image--question--mask triplets end-to-end, so constructing a benchmark that jointly evaluates answer correctness and pixel-level evidence grounding requires rigorous human control at every stage of the pipeline. \ours thus comprises 1,157 carefully curated samples spanning six task types, including \emph{identification}, \emph{attribute}, \emph{OCR}, \emph{spatial}, \emph{counting}, and \emph{comparison} and six visual domains, including \emph{indoor}, \emph{outdoor}, \emph{autonomous driving}, \emph{robotics}, \emph{science}, and \emph{mathematics}. These tasks cover the spectrum from direct perceptual recognition to compositional and relational reasoning in diverse real-world settings. These tasks and domains are illustrated in Figure~\ref{fig:teaser}. Notably, \ours includes hallucination-aware samples, where questions are unanswerable or refer to absent entities, requiring models to correctly identify the absence of valid visual evidence.

We formulate evaluation as a \emph{joint correctness} problem: a prediction is considered fully correct only when both the textual answer and the corresponding evidence mask are correct. To support this setting, we introduce \textsc{Grove} (\textbf{GRO}unded e\textbf{V}idence \textbf{E}valuation), a unified metric that jointly measures answer correctness and grounding fidelity by computing a per-sample geometric mean of smoothed text and mask scores, ensuring that both dimensions are satisfied simultaneously. \textsc{Grove} is designed around two key desiderata: (1)~\emph{joint sensitivity}, penalizing failures in either modality, and (2)~\emph{graceful degradation}, preserving signal under partial correctness, while remaining applicable across diverse model classes and evaluation settings.

Our main contributions are as follows:
\begin{itemize}
    \item We introduce \ours, a comprehensive benchmark for the joint evaluation of free-form answers and pixel-level supporting evidence across six tasks and six domains, with explicit hallucination-aware scenarios.
    
    \item We propose \textsc{Grove}, a unified evaluation metric that jointly measures answer correctness and grounding fidelity by computing a per-sample geometric mean of smoothed text and mask scores, ensuring that both dimensions are satisfied simultaneously.

    
    \item We conduct comprehensive experiments across state-of-the-art baselines on \ours, demonstrating that even the strongest current models struggle to align answers with correct visual evidence.
    
\end{itemize}

\section{Related work}

Early benchmarks for vision-language understanding focused on isolated capabilities, limiting their ability to evaluate grounded reasoning. VQA benchmarks have evolved from basic visual recognition~\cite{antol2015vqa,goyal2017vqa2} toward compositional and multi-hop 
reasoning~\cite{yue2025mmmupro, fu2025mme}, with subsequent efforts expanding coverage to specialized domains such as autonomous driving~\cite{marcu2024lingoqa,sima2024drivelm} and mathematics~\cite{lumathvista}, as well as to more challenging tasks such as hallucination mitigation~\cite{guan2024hallusionbench,zhang2025robust}. Despite this breadth, VQA benchmarks evaluate correctness at the textual level and do not require models to provide explicit visual evidence supporting their answers. 
A complementary line of work focuses on grounding through segmentation. Referring expression datasets~\cite{yu2016modelingrefcoco, mao2016generation, rasheed2024glamm, wu2024sesame} require models to localize objects described by natural language. More recent benchmarks extend this to reasoning-centric settings~\cite{lai2024lisa, 
li2025counterfactual, sahoo2026converseg}, introducing increasingly complex queries, counterfactual scenarios, and ambiguity. However, these benchmarks continue to treat segmentation masks as the primary prediction target. Despite their individual advances, the above benchmarks evaluate either \emph{what} models answer or \emph{where} they localize, but do not explicitly require answers to be grounded in visual evidence.

More recent benchmarks attempt to bridge the gap between VQA and grounded localization, yet critical limitations persist across three dimensions: grounding role, reasoning depth, and evaluation scope. Early efforts, such as VizWiz-Grounding~\cite{chen2022vizwizgrounding} and V$^*$Bench~\cite{wu2024vstar}, pair questions with visual localization, but treat grounding as a referring mechanism, identifying \emph{where} an answer object is located, rather than as evidence for \emph{why} an answer is correct. The works most closely related to ours are VRT-Bench~\cite{yuan2025vrt} and TreeBench~\cite{wang2026treebench}, which move toward evidence-based grounding. VRT-Bench associates reasoning steps with segmentation masks as supporting evidence, while TreeBench introduces structured reasoning with traceable bounding box signals. However, as summarized in Table~\ref{t:related_work}, both still evaluate text and grounding independently, rather than enforcing consistency between them. They remain confined to a single domain and do not explicitly assess hallucination robustness, leaving a critical gap in comprehensive grounded evaluation. \ours is designed to bridge this gap by introducing a unified evaluation metric that jointly assesses answer correctness and pixel-level grounding fidelity. The proposed metric enables a rigorous evaluation of whether model predictions are both factually accurate and visually supported. Spanning six tasks and six domains, \ours covers a spectrum from basic perception to high-level reasoning across diverse real-world settings, while explicitly accounting for model hallucination. It is also worth noting that, beyond the evaluation-set sizes reported in Table~\ref{t:related_work}, some existing benchmarks provide associated training data for their corresponding tasks (e.g., 239 samples in ReasonSeg, approximately 61K in ConverSeg, and over 200K in GranD), whereas \ours is designed as an evaluation-only benchmark.

\begin{table*}[t]
\caption{Comparison of recent benchmarks for grounded VQA and segmentation. MC: Multiple Choice; Acc.: Text Accuracy.
}
\label{t:related_work}
\centering
\resizebox{\textwidth}{!}{
\begin{tabular}{l|cccccccc}
\toprule
\textbf{Name} & \textbf{Test Set Size} & \textbf{VQA / Q type} & \textbf{Grounding} & \textbf{Halluc.} & \textbf{\#Tasks} & \textbf{\#Domains} & \textbf{Annot.} & \textbf{Metric} \\
\midrule

ReasonSeg~\cite{lai2024lisa} & 779 & ${\color{red}\times}$ & mask (reasoning) & ${\color{red}\times}$ & 1 & 1 &  {\color{pink}\faUser} {\color{teal}\faCogs} & gIoU \\

GranD~\cite{rasheed2024glamm} & 5000 & ${\color{red}\times}$ & mask (referring) &  ${\color{red}\times}$ & 1 & 1 & {\color{teal}\faCogs} & mIoU  \\

FP-RefCOCO~\cite{wu2024sesame} & 1500 & ${\color{red}\times}$ & mask (referring)  &  {\color{blue}\checkmark} & 1 & 1 & {\color{teal}\faCogs} & cIoU  \\

HalluSegBench~\cite{li2025counterfactual} & 399 & ${\color{red}\times}$ & mask (reasoning) & {\color{blue}\checkmark} & 1 & 1 &  {\color{pink}\faUser} {\color{teal}\faCogs}  & CMS, mIoU \\

ConverSeg~\cite{sahoo2026converseg} & 1687 & ${\color{red}\times}$ & mask (reasoning) & {\color{blue}\checkmark} & 4 & 1 &  {\color{pink}\faUser} {\color{teal}\faCogs}  & gIoU \\

V$^*$Bench~\cite{wu2024vstar} & 191 & {\color{blue}\checkmark}/ MC & bbox (referring) & ${\color{red}\times}$ & 2 & 1 &  {\color{pink}\faUser} & Acc. \\

VizWiz-VQA-Gr~\cite{chen2022vizwizgrounding} & 2373 & {\color{blue}\checkmark}/ free-form & bbox (referring)  & ${\color{red}\times}$ & 1 & 1 & {\color{pink}\faUser} & mIoU \\

TreeBench~\cite{wang2026treebench}  & 405 & {\color{blue}\checkmark}/ MC & bbox (evidence) & ${\color{red}\times}$ & 10 & 1 & {\color{pink}\faUser} {\color{teal}\faCogs} & Acc., mIoU \\ 

VRT-Bench~\cite{yuan2025vrt} & 304 & {\color{blue}\checkmark}/ free-form & mask (evidence) & ${\color{red}\times}$ & 4 & 1 & {\color{pink}\faUser} {\color{teal}\faCogs} & Acc., mIoU  \\

\midrule
\textbf{\ours (ours)} & 1157 & {\color{blue}\checkmark}/ free-form & mask (evidence) &{\color{blue}\checkmark} & 6 & \textbf{6} & {\color{pink}\faUser} {\color{teal}\faCogs} & Joint (\textbf{\metric}) \\
\bottomrule
\end{tabular}
}
\vspace{-1em}
\end{table*}

\section{\ours Benchmark} \label{benchmark}

We introduce \ours, a comprehensive benchmark for the joint evaluation of answer correctness and pixel-level evidence grounding in VQA. Unlike prior benchmarks that assess text and grounding independently, \ours requires models to produce a free-form answer alongside a segmentation mask that serves as explicit visual evidence supporting that answer. A prediction is considered correct only when both the answer and the corresponding mask are valid, enabling rigorous assessment of whether model outputs are both accurate and visually grounded. 

\subsection{Tasks and Domains}

\ours comprises 1,157 curated samples covering six task types and six visual domains, including 314 hallucination samples ($\approx$27\%) specifically designed to probe robustness against visually misleading or unanswerable queries. Figure~\ref{fig:teaser} illustrates representative examples from all task types. Detailed definitions are provided in Appendix~\ref{app:task}.

\textbf{Tasks.} \ours's six task types span a spectrum from direct visual perception to higher-order reasoning, reflecting the diverse cognitive demands of grounded VQA. Perception-oriented tasks (\emph{identification}, \emph{attribute}, and \emph{OCR}) evaluate first-order visual understanding, where correct answers depend on accurately localizing a target region and directly interpreting its properties.
Reasoning-oriented tasks (\emph{spatial}, \emph{counting}, and \emph{comparison}) require higher-order inference over multiple regions, involving relational, enumerative, or comparative reasoning beyond direct recognition.

\textbf{Domains.} \ours's six visual domains are designed to capture the diversity of real-world settings in which grounded visual reasoning is required. Together, they span a broad spectrum of visual characteristics and reasoning demands, from natural scenes to structured diagrams, and from passive perception to embodied interaction, enabling comprehensive evaluation of model robustness across diverse scenarios: (1)~\emph{indoor} and (2)~\emph{outdoor} environments represent everyday scenes with varying object density, lighting, and contextual complexity, testing robust grounded understanding in general settings; (3)~\emph{autonomous driving} introduces safety-critical scenarios with dynamic agents and structured road semantics; (4)~\emph{robotics} focuses on embodied interaction and manipulation-centric reasoning, where tasks depend on accurately localizing objects and reasoning about their spatial configuration and affordances; (5)~\emph{science} includes domain-specific visual content such as diagrams, charts, and biological or physical schematics, requiring grounding over structured representations that combine perceptual and domain knowledge; and (6)~\emph{math} emphasizes geometric interpretation and diagram-based reasoning.

\subsection{Benchmark Construction}
Constructing a benchmark that jointly evaluates answer correctness and pixel-level evidence grounding requires rigorous control at every stage of the pipeline. A key challenge is that current state-of-the-art MLLMs are insufficiently reliable to autonomously generate complex image--question--mask triplets end-to-end. In particular, MLLMs can introduce systematic errors in QA generation for certain tasks (see Appendix~\ref{app:failure}), while reasoning segmentation models often fail to produce accurate masks when prompted with complex or domain-specific queries, rendering fully automated construction unreliable. Accordingly, \ours is constructed using a systematic multi-stage pipeline that combines domain-appropriate image sourcing, hybrid mask generation, LLM-assisted QA generation, and multi-round human verification, ensuring that each sample satisfies strict standards of visual quality, answer correctness, and grounding fidelity.

\textbf{Image Collection and Generation.}
We sample approximately 300 images per domain from four existing datasets and generate 200 synthetic images for the mathematics domain, with deliberate emphasis on visual complexity and reasoning challenge. Images from existing datasets are initially selected through random sampling, followed by a screening step to ensure their suitability for at least one of the benchmark's reasoning tasks (e.g., legible text for OCR or sufficient object diversity for counting and comparison). For \textit{indoor} and \textit{outdoor} scenes, images are sampled from SA-1B~\cite{kirillov2023segment}, which offers high-resolution, real-world scenes with a large number of small and varied objects, making it particularly suitable for evaluating visually grounded reasoning across everyday environments. \textit{Autonomous driving} images are sampled from NuScenes~\cite{caesar2020nuscenes}, capturing complex multi-agent scenarios with safety-critical objects and structured road semantics. \textit{Robotics} images are sampled from the DROID dataset~\cite{khazatsky2024droid}, focusing on manipulation and interaction scenes that demand fine-grained spatial and object-level understanding. Due to storage constraints, DROID frames are randomly sampled from its first 300 episodes. \textit{Science} images are sampled from ScienceQA~\cite{saikh2022scienceqa}, covering domain-specific diagrams, charts, and illustrations that require grounding in structured visual representations. For \textit{mathematics}, images are generated using custom scripts, enabling precise control over geometric configurations, symbolic layouts, and ground-truth answers.

\textbf{Mask Generation and Extraction.}
Segmentation masks are constructed using domain-specific strategies tailored to the characteristics of each data source. For \textit{indoor} and \textit{outdoor} domains, masks are directly extracted from existing SA-1B annotations, providing high-quality pixel-level segmentations. For \textit{mathematics}, masks are generated programmatically alongside the images, ensuring exact correspondence between visual content and evidence regions. For \textit{autonomous driving}, \textit{robotics}, and \textit{science} domains, masks are obtained through a combination of automated segmentation using SAM3~\cite{carion2025sam} and human annotation. In practice, SAM3 often fails to reliably isolate specific instances from text prompts and struggles with sub-structure recognition in specialized domains (\eg, distinguishing fine-grained biological components such as organelles). To address these limitations, annotators manually draw or refine masks to accurately delineate the intended evidence regions.

\textbf{Question-Answer Generation.}
For each image, we use two MLLMs, GPT-5.2~\cite{openai2025gpt52} and Gemini 3 Pro~\cite{google2025gemini3}, to independently generate five candidate QA pairs each, conditioned on the image, task type, and a structured prompt template crafted to ensure grounding-aware question complexity and precise visual-semantic correspondence (see Appendix~\ref{app:data-construct}). To evaluate models tendency to hallucinate, we also include an adversarial objective: models must generate three negative candidates involving non-existent entities, invalid spatial relations, or multi-hop reasoning anchored in absent attributes. Human annotators then select the most semantically precise candidate from the resulting pool of ten/six, or manually compose or refine QA for task types when automated generation consistently fails, most notably for reasoning-intensive tasks such as \textit{counting}.

\textbf{Quality Control} is applied at three stages of the pipeline. Following mask generation, annotators verify image and mask quality, ensuring that each mask accurately captures the intended evidence region. Following QA generation, annotators evaluate each candidate \wrt the intended reasoning skill, precision of visual grounding, and clarity of answerability, selecting the most semantically precise candidate or performing manual refinement when necessary. A final cross-validation round by independent annotators ensures the consistency of the complete question–answer–mask triplet (see Appendix~\ref{app:ver}). The final benchmark comprises 1,157 samples after rigorous quality filtering.

\subsection{Statistics}
\begin{figure*}[t]
    \centering
    \begin{subfigure}[b]{0.29\textwidth}
        \centering
        \includegraphics[width=\textwidth]{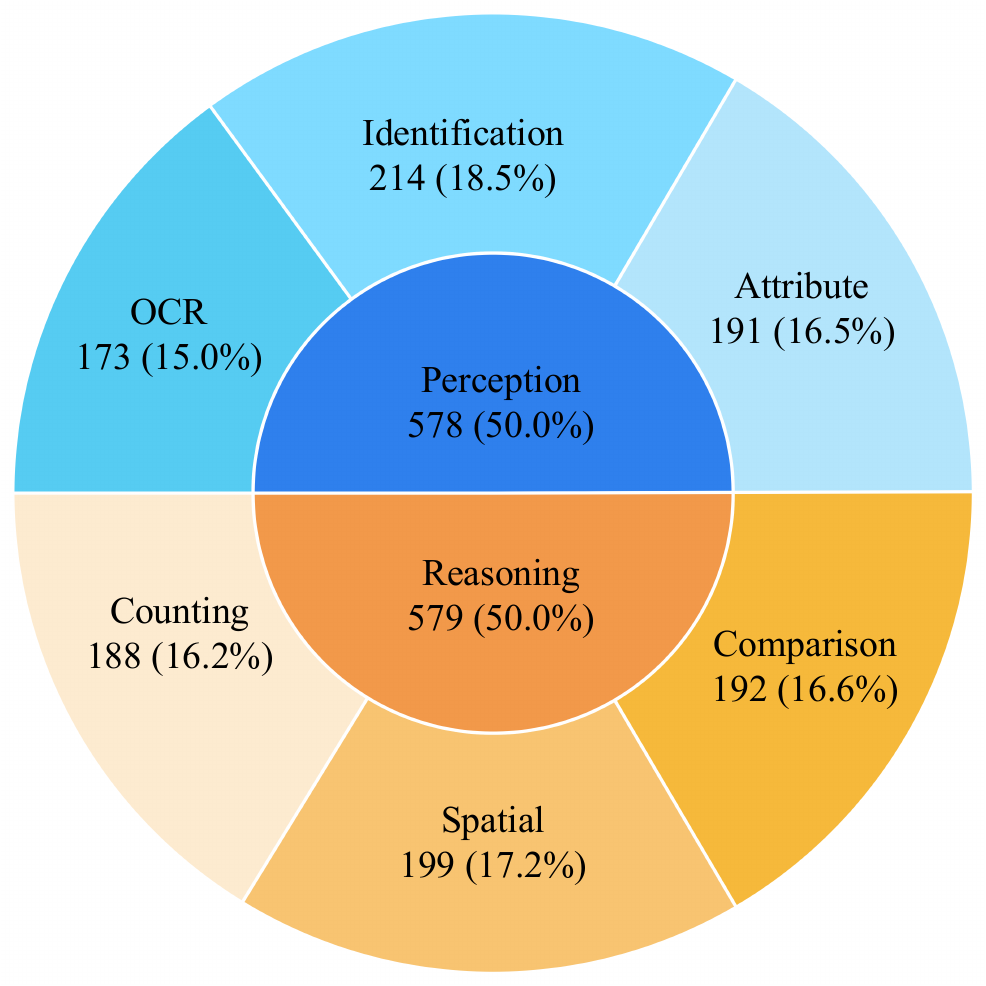}
        \caption{Task distribution.}
        \label{fig:stats_tasks}
    \end{subfigure}
    \hfill
    \begin{subfigure}[b]{0.29\textwidth}
        \centering
        \includegraphics[width=\textwidth]{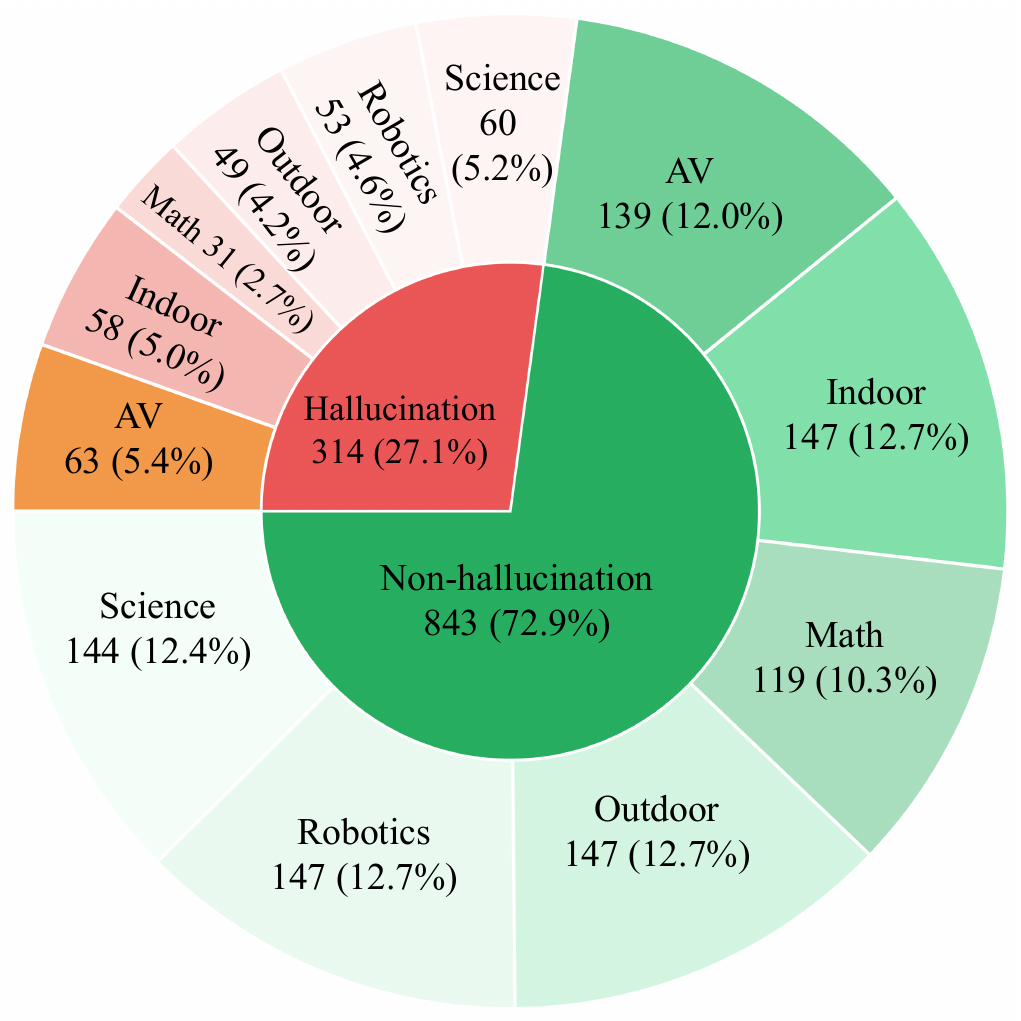}
        \caption{Domain distribution.}
        \label{fig:stats_domain}
    \end{subfigure}
    \hfill
    \begin{subfigure}[b]{0.38\textwidth}
        \centering
        \includegraphics[width=\textwidth]{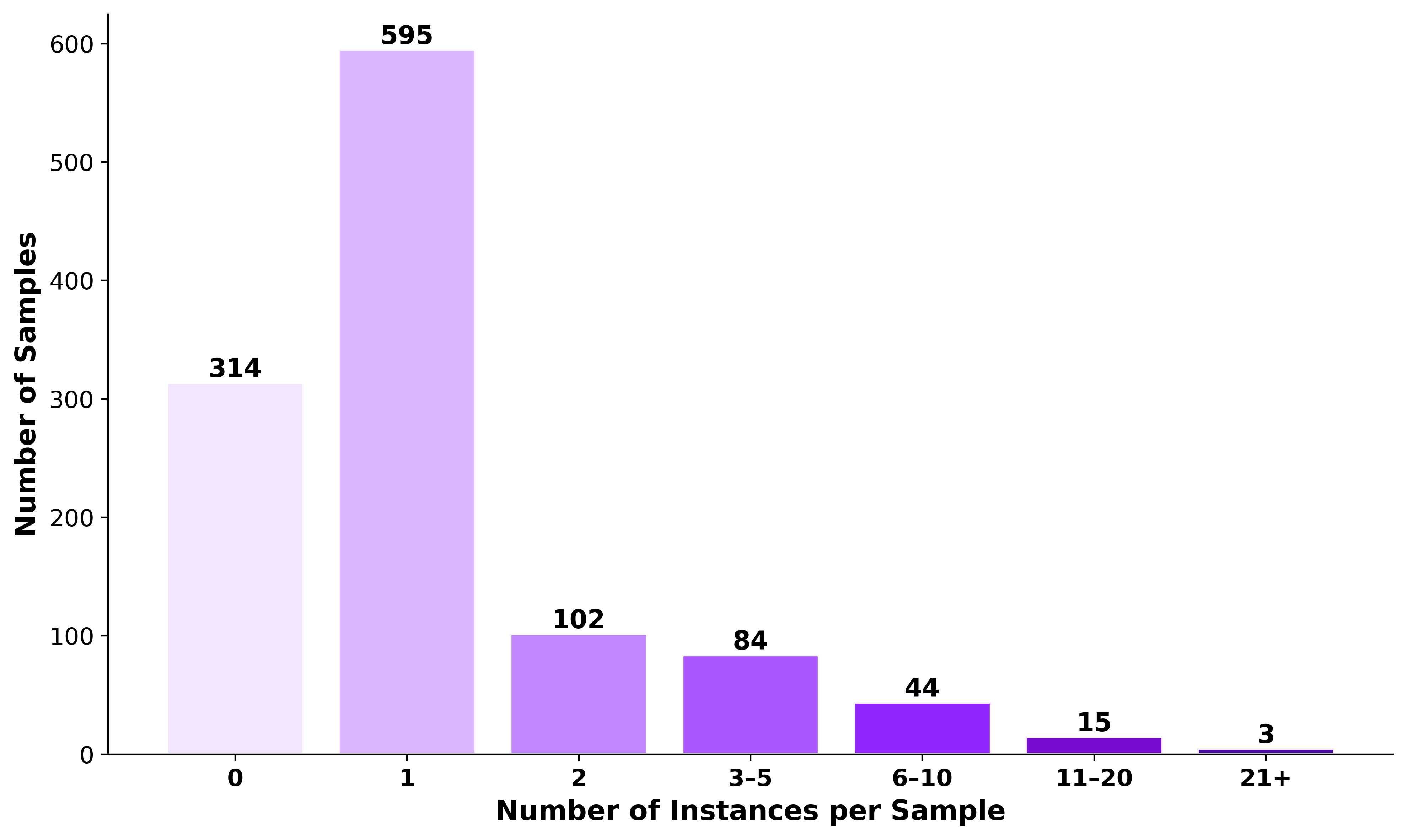}
        \caption{Mask multiplicity}
        \label{fig:stats_instances}
    \end{subfigure}
\caption{\ours dataset statistics. The benchmark is balanced across 
task types (\subref{fig:stats_tasks}) and across domains for 
hallucination and non-hallucination samples (\subref{fig:stats_domain}). 
Mask multiplicity follows a long-tailed distribution driven primarily 
by \textit{counting} tasks, where the zero-instance samples 
correspond to hallucination cases (\subref{fig:stats_instances}).}
    \label{fig:stats}
\end{figure*}

The distribution of \ours's task types and domains is illustrated in Figure~\ref{fig:stats}.
Task types are balanced between perception and reasoning categories, with individual tasks ranging from 173 (OCR, 15.0\%) to 214 (Identification, 18.5\%) samples. This balance is deliberate: unlike benchmarks that emphasize higher-order reasoning, \ours evaluates grounding fidelity across the full spectrum of visual cognition, reflecting the requirement that evidence grounding must remain reliable regardless of task complexity. Figure~\ref{fig:stats_domain} shows that approximately 27.1\% of samples (314) consist of hallucination cases where no valid grounding exists, explicitly probing model robustness to misleading or unanswerable queries. The remaining 72.9\% (843) correspond to grounded scenarios with at least one annotated evidence mask. Domain coverage is similarly balanced across all six visual settings, ensuring that evaluation is not confounded by domain bias.

Mask multiplicity follows a long-tailed distribution (Figure~\ref{fig:stats_instances}), where the 314 zero-instance samples correspond to hallucination cases in which the queried entity is absent by design. Among the 843 grounded samples, 70.6\% contain a single mask instance, reflecting the prevalence of precise, localized evidence grounding, while 29.4\% contain two or more instances, with a long tail extending to a maximum of 57 masks per sample. Overall, single- and dual-instance samples account for 82.7\% of grounded samples, indicating a balance between grounding precision and compositional complexity. Multi-instance scenarios, driven primarily by \textit{counting} tasks (mean of 3.89 masks per sample), further stress-test instance-level grounding in dense scenes.

\subsection{Evaluation Metric}

We introduce \metric (\textbf{GR}ounded e\textbf{V}idence \textbf{E}valuation), a unified metric designed around two key desiderata: (1)~\textit{joint sensitivity}, where neither text nor mask correctness alone is sufficient for a high score, and (2)~\textit{graceful degradation}, where partial correctness along one axis contributes a meaningful signal rather than collapsing to zero. 

\textbf{Answer Score} ($S_a \in \{0, 1\}$) is a binary correctness signal obtained via an LLM-as-judge protocol, which has shown to be correlated strongly with human judgment for free-form answer evaluation~\cite{zheng2023judging} (also see  Section~\ref{discussion}). Given the question, ground-truth answer, and predicted answer, the judge model returns a binary verdict based on semantic correctness rather than surface-level string matching. We use Qwen 2.5-14B~\cite{yang2025qwen3} as the judge model and provide the prompt template in Appendix~\ref{app:judge}.

\textbf{Mask Score} ($S_m \in [0, 1]$) measures the fidelity of predicted segmentation masks as visual evidence. To handle single-instance, multi-instance, and hallucination samples uniformly, $S_m$ is defined as:
\begin{equation}
S_m(P, G) = \begin{cases} 
1 & \text{if } G = \emptyset \wedge P = \emptyset \\ 
0 & \text{if } G = \emptyset \wedge P \neq \emptyset \\ 
\dfrac{1}{\max(|P|, |G|)} \displaystyle\sum_{(p,g) \in \mathcal{B}} 
\text{IoU}(p, g) & \text{otherwise} 
\end{cases}
\end{equation}
\noindent where $P$ is the set of predicted masks, $G$ is the set of ground-truth masks, and $\mathcal{B}$ is the bipartite matching between $P$ and $G$ obtained via the Hungarian algorithm~\cite{kuhn1955hungarian}. The first case rewards correct predictions of no mask when none exists, corresponding to hallucination samples where the queried entity is absent from the scene. The second case penalizes spurious mask predictions when the ground truth is empty, reflecting a hallucination failure. The third case computes the mean IoU over the optimal matching, normalized by $\max(|P|, |G|)$ to penalize both missed and spurious predictions.

\textbf{Joint Score (\metric).}
Directly multiplying $S_a$ and $S_m$ yields a degenerate metric: any 
sample with $S_a = 0$ collapses to zero regardless of mask quality, 
potentially discarding meaningful grounding signals. To address this, we apply $\epsilon$-floor smoothing to both scores before computing the joint metric:
\begin{align}
    S_a' &= \max(S_a,\ \epsilon) \\
    S_m' &= \max(S_m,\ \epsilon)
\end{align}
where $\epsilon$ prevents score collapse while preserving sensitivity to partial correctness. We empirically choose a value of $\epsilon = 0.1$, which yields good score discriminability while balancing our desiderata of joint sensitivity and graceful degradation (see section~\ref{discussion} and Appendix~\ref{app:epsilon}).
The per-sample score is then defined as the geometric mean 
of the smoothed scores:
\begin{equation}
    \mathcal{S} = \sqrt{S_a' \cdot S_m'} \label{eq:per-sample}
\end{equation}
The geometric mean enforces joint competence by penalizing imbalance between the two axes~\cite{steele2004cauchy}, ensuring that strong performance on one axis cannot compensate for failure on the other.

\textbf{Benchmark Score.}
The \metric score with respect to $N$ samples is computed as the mean of their per-sample scores:
\begin{equation}
    \textsc{Grove} = \frac{1}{N} \sum_{i=1}^{N} \mathcal{S}_i
    \label{eq:grove}
\end{equation}

where $\mathcal{S}_i$ is the smoothed per-sample score~\eqref{eq:per-sample}.

\section{Experiments} \label{experiments}

We evaluate various models spanning dedicated grounding and reasoning segmentation architectures, as well as pipeline approaches pairing frontier MLLMs with SAM3~\cite{carion2025sam}. All models are evaluated zero-shot without fine-tuning on \ours and are provided with standardized output format instructions (Appendix~\ref{app:output-format}). All scores are reported as percentages. In addition to \metric scores, we report overall text accuracy ($\mathcal{T}$) and overall mIoU ($\mathcal{M}$) for reference. These metrics are not directly comparable with \metric, as the latter applies $\epsilon$-flooring and aggregates text and mask scores jointly at the sample level, rather than averaging each component independently. Consequently, \metric may exceed either marginal score, as further discussed in Appendix~\ref{app:epsilon}. All evaluations are conducted on 4 NVIDIA GeForce RTX 4090 GPUs.

\subsection{Results} \label{sec:results}

Table~\ref{tab:task_results} presents the overall performance of our chosen models using \ours, along with a breakdown by task type. All models achieve low \metric scores, with the strongest model, GPT-5.4-Thinking + SAM3 (GPT-5.4-T + SAM3), reaching only 45.53, highlighting the benchmark’s rigor in requiring simultaneous answer correctness and pixel-level evidence grounding. Among grounding-aware and reasoning segmentation models, VRT-RL achieves the best performance (29.02), followed by Sa2VA (25.74). VRT-RL's performance is consistent with its reinforcement learning post-training reward design, while reasoning-segmentation models such as LISA, GLaMM, and LaSagnA exhibit relatively low answer correctness, reflecting their primary optimization for visual grounding rather than open-ended question answering. Hybrid pipelines consistently outperform grounding models, with GPT-5.4-T + SAM3 achieving the highest overall score, followed by Qwen3-VL-4B-I+ SAM3.

Crucially, a modality gap persists: while models may achieve relatively high text accuracy and mIoU mask scores, the joint \metric remains low, indicating that correct answers and accurate evidence often occur on different samples. For instance, Gemini 3 + SAM3 achieves the highest answer correctness (62.32), yet obtains the lowest \metric score among hybrid pipelines (39.63). Similarly, Qwen3-VL-32B-I + SAM3 improves answer correctness over its 4B counterpart (57.65 vs. 53.15), but exhibits lower grounding quality (36.02 vs. 39.91), resulting in no improvement in overall \metric. These results demonstrate that high aggregate text and mask scores do not necessarily imply that both properties are satisfied on the same samples, motivating the per-sample joint evaluation of \metric. This discrepancy also reflects architectural differences: LISA-style models generate a dedicated [SEG] token whose hidden-state embedding conditions the mask decoder, potentially propagating answer hallucinations into grounding predictions. Sa2VA additionally incorporates visual features, whereas hybrid pipelines condition SAM3 on MLLM-generated textual prompts, coupling answer generation and grounding through different mechanisms. Across tasks, spatial reasoning and OCR emerge as the most challenging, reflecting their higher compositional and grounding demands.

\begin{table*}[t]
\caption{Per-task and overall results using \ours. We report \metric scores along with overall text accuracy ($\mathcal{T}$) and overall mIoU ($\mathcal{M}$) scores. $^\dagger$ \texttt{R-Sa2VA-Qwen3VL-4B-RL} checkpoint.}
\centering
\resizebox{\textwidth}{!}{
\begin{tabular}{l | c c c c c c | c c | c}
\toprule
& \multicolumn{6}{c|}{\textbf{Per-task \metric scores}} & \multicolumn{3}{c}{\textbf{Overall scores}} \\
\cmidrule{2-7} \cmidrule{8-10}
\textbf{Model} & \textbf{Identification} & \textbf{Attribute} & \textbf{OCR} & \textbf{Spatial} & \textbf{Counting} & \textbf{Comparison} & \textbf{$\mathcal{T}$} & \textbf{$\mathcal{M}$} & \metric \\

\midrule
LISA-7B~\cite{lai2024lisa}       & 15.16 & 16.92 & 12.44 & 15.29 & 16.56 & 16.23 & 7.26  & 17.81 & 15.47 \\
SESAME-7B~\cite{wu2024sesame}    & 17.90 & 17.14 & 13.62 & 18.48 & 17.60 & 16.90 & 3.20  & 23.67 & 17.02 \\
GLaMM-7B~\cite{rasheed2024glamm} & 14.86 & 12.89 & 12.34 & 14.19 & 11.84 & 14.70 & 1.38  & 14.69 & 13.53 \\
LaSagnA-7B~\cite{wei2024lasagna} & 13.31 & 12.03 & 11.04 & 13.23 & 11.14 & 13.05 & 0.43  & 10.74 & 12.35 \\
UniPixel-7B~\cite{liuunipixel}   & 22.24 & 28.29 & 22.19 & 19.90 & 15.36 & 26.34 & 21.26 & 20.43 & 22.39 \\
Sa2VA-8B~\cite{yuan2025sa2va}    & 26.79 & 30.44 & 25.36 & 22.87 & 20.46 & 28.36 & 29.04 & 21.01 & 25.74 \\
VRT-RL$^\dagger$~\cite{yuan2025vrt} & 34.92 & 28.74 & 27.43 & 24.44 & 29.94 & 28.01 & 36.30 & 24.25 & 29.02 \\
TreeVGR-7B~\cite{wang2026treebench} & 18.28 & 18.55 & 16.42 & 17.90 & 20.25 & 22.65 & 16.08 & 19.66 & 19.03 \\
Uground-7B~\cite{qian2025uground}   & 10.95 & 11.69 & 11.15 & 11.57 & 12.15 & 13.22 & 0.95  & 7.20  & 11.78 \\
\midrule
Qwen3-VL-4B-I + SAM3   & 50.36 & 47.46 & 40.96 & 36.32 & 45.52 & 38.43 & 53.15 & 39.91 & 43.30 \\
Qwen3-VL-32B-I + SAM3  & 48.41 & 46.37 & 42.14 & 35.91 & 45.72 & 37.41 & 57.65 & 36.02 & 42.72 \\
Gemini 3 + SAM3         & 43.10 & 40.11 & 42.55 & 33.53 & 42.81 & 35.84 & 62.32 & 34.99 & 39.63 \\
GPT-5.4 + SAM3          & 48.22 & 40.43 & 41.33 & 35.62 & 47.06 & 35.86 & 53.50 & 39.26 & 41.50 \\
GPT-5.4-T + SAM3        & 53.62 & 44.73 & 44.16 & 38.86 & 52.21 & 38.93 & 61.02 & 41.98 & 45.53 \\

\bottomrule
\end{tabular}
}

\label{tab:task_results}
\end{table*}

\begin{table*}[t]
\caption{Per-domain and hallucination results using \ours. We report \metric scores along with overall text accuracy ($\mathcal{T}$) and overall mIoU ($\mathcal{M}$) for hallucination and non-hallucination subsets. $^\dagger$ denotes \texttt{R-Sa2VA-Qwen3VL-4B-RL} checkpoint.}

\centering
\resizebox{\textwidth}{!}{
\begin{tabular}{l | c c c c c c | c c c | c c c}
\toprule
& \multicolumn{6}{c|}{\textbf{Per-domain \metric scores}} & \multicolumn{3}{c|}{\textbf{Hallucination scores}} & \multicolumn{3}{c}{\textbf{Non-Hallucination scores}} \\
\cmidrule{2-7} \cmidrule{8-10} \cmidrule{11-13}
\textbf{Model} 
& \textbf{Indoor} & \textbf{Outdoor} & \textbf{Math} & \textbf{AV} & \textbf{Rob.} & \textbf{Sci.} 
& $\mathcal{T}$ & $\mathcal{M}$ & \metric 
& $\mathcal{T}$ & $\mathcal{M}$ & \metric \\
\midrule
LISA-7B~\cite{lai2024lisa}           & 16.01 & 16.41 & 12.15 & 16.50 & 15.97 & 14.95 & 1.59  & 40.45 & 19.39 & 9.37  & 9.38  & 14.01 \\
SESAME-7B~\cite{wu2024sesame}        & 17.20 & 17.68 & 11.05 & 20.77 & 15.92 & 17.97 & 9.87  & 61.46 & 30.04 & 0.71  & 9.59  & 12.17 \\
GLaMM-7B~\cite{rasheed2024glamm}     & 12.82 & 13.75 & 11.13 & 15.70 & 13.43 & 13.72 & 0.32  & 4.14  & 10.96 & 1.78  & 18.62 & 14.48 \\
LaSagnA-7B~\cite{wei2024lasagna}     & 11.31 & 12.18 & 11.07 & 14.05 & 12.09 & 13.07 & 1.27  & 0.00  & 10.28 & 0.12  & 14.74 & 13.12 \\
UniPixel-7B~\cite{liuunipixel}       & 21.76 & 23.20 & 12.90 & 28.87 & 21.02 & 24.15 & 12.74 & 0.00  & 12.75 & 24.44 & 28.04 & 25.98 \\
Sa2VA-8B~\cite{yuan2025sa2va}        & 22.64 & 24.03 & 14.87 & 36.87 & 25.10 & 28.09 & 7.32  & 0.00  & 11.58 & 37.13 & 28.84 & 31.01 \\
VRT-RL$^\dagger$~\cite{yuan2025vrt}  & 29.61 & 33.06 & 24.68 & 30.76 & 27.74 & 27.27 & 9.24  & 24.84 & 20.79 & 46.38 & 24.04 & 32.09 \\
TreeVGR-7B~\cite{wang2026treebench}  & 18.41 & 20.45 & 12.70 & 23.51 & 19.76 & 17.78 & 26.43 & 24.52 & 22.66 & 12.22 & 17.85 & 17.68 \\
Uground-7B~\cite{qian2025uground}    & 11.32 & 10.74 & 10.78 & 11.79 & 13.40 & 12.38 & 2.55  & 14.01 & 14.03 & 0.36  & 4.66  & 10.94 \\
\midrule
Qwen3-VL-4B-I + SAM3   & 43.64 & 46.84 & 27.37 & 59.39 & 41.78 & 36.80 & 55.73 & 70.70 & 60.72 & 52.19 & 28.44 & 36.81 \\
Qwen3-VL-32B-I + SAM3  & 44.27 & 47.63 & 23.23 & 58.40 & 40.09 & 37.85 & 61.78 & 55.73 & 56.41 & 56.11 & 28.68 & 37.63 \\
Gemini 3 + SAM3         & 39.37 & 39.40 & 31.40 & 48.42 & 37.15 & 39.87 & 40.76 & 92.04 & 56.14 & 70.34 & 13.75 & 33.48 \\
GPT-5.4 + SAM3          & 42.06 & 46.05 & 28.99 & 53.43 & 39.87 & 35.54 & 25.16 & 55.73 & 36.72 & 64.06 & 33.12 & 43.28 \\
GPT-5.4-T + SAM3        & 45.39 & 45.92 & 41.51 & 53.18 & 47.01 & 39.24 & 39.17 & 67.20 & 47.89 & 69.16 & 32.59 & 44.65 \\

\bottomrule
\end{tabular}
}

\label{tab:domain_results}
\end{table*}

Table~\ref{tab:domain_results} reports per-domain and hallucination breakdown results. Across domains, \emph{Math} is consistently the most challenging for all model families, likely due to the abstraction required to ground symbolic reasoning in pixels. Hallucination samples specifically assess whether models can recognize the absence of queried entities and produce both an appropriate textual response and an empty evidence mask. Several segmentation models, including UniPixel, Sa2VA, and LaSagnA, continue to predict masks for nonexistent entities, yielding zero hallucination mask scores. In contrast, SESAME and TreeVGR demonstrate stronger hallucination-aware performance, with SESAME benefiting from explicit non-existent-object supervision during training. To further disentangle textual recognition from empty-mask inference, we jointly analyze both predictions across hallucination samples. Across all models, only 14.7\% of predictions correctly recognize the hallucination and produce an empty mask, while 6.3\% correctly recognize the hallucination but predict an incorrect mask. Notably, 21.8\% predict an empty mask despite an incorrect textual response, indicating that empty-mask prediction alone does not imply successful hallucination recognition. Among textually correct predictions, approximately 70\% also produce the correct empty mask. These findings demonstrate that hallucination-aware performance reflects both the ability to recognize unsupported queries and the consistency of the corresponding grounding predictions, rather than empty-mask inference alone. Detailed per-task and per-domain text and mask score breakdowns are provided in Appendix~\ref{app:result}.

\subsection{Qualitative Results}

\begin{figure*}[t]
    \centering
    \includegraphics[width=\textwidth]{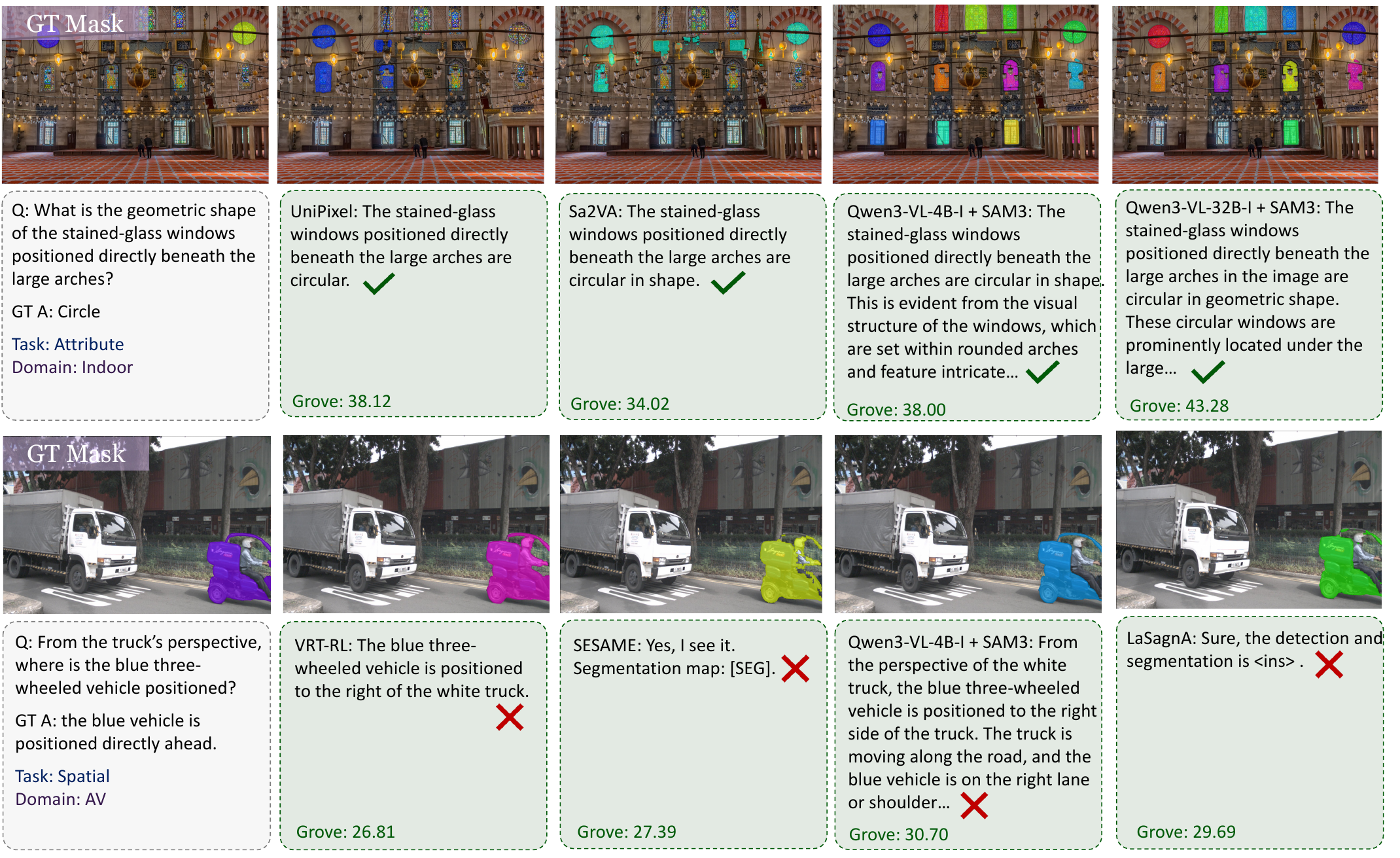}
    \vspace{-1em}
    \caption{Illustration of the modality gap on \ours. Top: correct answers (\textcolor{green!50!black}{$S_a = 1$}) but inaccurate evidence (\textcolor{red}{$S_m < 0.2$}). Bottom: accurate evidence (\textcolor{green!50!black}{$S_m > 0.7$}) but incorrect answers (\textcolor{red}{$S_a = 0$}). In both cases, partial correctness in a single modality yields low \metric scores, highlighting the need for joint evaluation.
    }
    \label{fig:QR}
\end{figure*}

Figure~\ref{fig:QR} illustrates the modality gap across two representative failure modes. In the top row, models spanning grounding-specific (UniPixel, Sa2VA) and hybrid pipelines (Qwen3-VL-4B, Qwen3-VL-32B) correctly identify that the stained-glass windows are circular ($S_a = 1$), yet segment irrelevant regions, resulting in low \metric scores (34–43) despite perfect text accuracy. In the bottom row, all shown models accurately localize the blue vehicle ($S_m > 0.7$) but misidentify its spatial relation to the truck, answering \emph{right} instead of \emph{ahead}, again yielding low \metric scores (27–31) despite strong grounding. Crucially, both failure modes remain invisible under single-modality evaluation: the top row would be deemed correct under text-only metrics, while the bottom row would score well under grounding-only metrics. A joint evaluation via \metric reveals these failures. Additional qualitative results, including further examples illustrating \metric behavior along with hallucination-aware cases are provided in Appendix~\ref{app:qual}.

\subsection{Discussion}\label{discussion}
\begin{wraptable}{l}{0.30\textwidth}
\vspace{-1.2em}
\centering
\caption{Agreement of various judges with human judgments.}
\resizebox{0.30\textwidth}{!}{
\begin{tabular}{lccc}
\toprule
Judge & Acc. & $\kappa$ & $F_1$ \\
\midrule
Exact Match   & 0.520 & 0.000 & 0.000 \\
Qwen2.5-14B   & 0.920 & 0.840 & 0.921 \\
GPT-5.4       & 0.932 & 0.864 & 0.931 \\
\bottomrule
\end{tabular}
}
\label{tab:judge_validation}
\vspace{-1em}
\end{wraptable}
\textbf{LLM-as-Judge reliability for free-form text correctness.} To validate our LLM judge, we sample $N=250$ responses balanced across tasks, domains, models, and answer correctness. Table~\ref{tab:judge_validation} reports accuracy, Cohen's $\kappa$~\cite{cohen1960coefficientkappa}, and $F_1$ for three evaluation methods against human judgments. Exact matching only achieves near-chance accuracy ($52.0\%$, barely above the 50\% random baseline for binary evaluation) and zero human agreement ($\kappa = 0.00$), confirming that lexical overlap is insufficient for evaluating free-form answers in \ours. In contrast, Qwen2.5-14B demonstrates strong agreement with human judgments ($\kappa = 0.840$), closely matching GPT-5.4 ($\kappa = 0.864$), validating its use as a reliable and cost-efficient binary correctness judge.

\textbf{Choice of aggregation function.} While several aggregation functions can combine answer correctness and grounding quality, they differ in how they penalize imbalanced predictions. The arithmetic mean permits linear compensation between the two components, whereas the geometric mean favors their joint satisfaction. The harmonic mean imposes a stronger imbalance penalty but is more sensitive to near-zero values and the choice of $\epsilon$, while the raw product compresses the score range. Importantly, recomputing per-sample joint scores for all 14 evaluated models using the geometric mean, harmonic mean, and raw product yields identical model rankings (Spearman's $\rho=1.0$, Kendall's $\tau=1.0$), indicating that our comparative conclusions are robust to the choice of these aggregation functions. Further details are provided in Appendix~\ref{app:aggreg}.

\textbf{Sensitivity of \metric to \texorpdfstring{$\epsilon$}{epsilon}.} \metric uses the flooring parameter $\epsilon$ to prevent score collapse while preserving sensitivity to sub-component variance. We adopt $\epsilon = 0.1$ as this value yields the greatest score discriminability without compromising joint sensitivity. To validate our choice, we performed a sensitivity analysis with values of $\epsilon \in \{0.01, 0.05, 0.1\}$. Using these values, we found that model rankings are preserved at the overall score level (Spearman’s rank correlation $\rho = 1.00$) and remain stable across task types and visual domains ($\rho \geq 0.971$ per task, $\rho \geq 0.974$ per domain), suggesting that \metric is robust to the choice of $\epsilon\in[0.01, 0.1]$. The complete results are given in~\Cref{app:tab_sensitivity_rho,app:tab_sensitivity_scores} in Appendix~\ref{app:epsilon}.

\begin{wrapfigure}{r}{0.30\textwidth}
\vspace{-1em}
\centering
\includegraphics[width=0.30\textwidth]{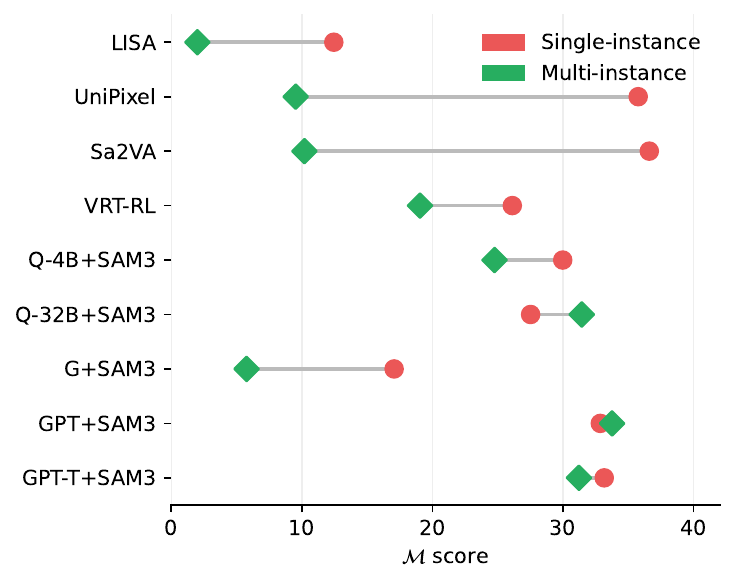}
\caption{Single- vs. multi-instance $\mathcal{M}$ scores. }
\label{fig:single_multi}
\vspace{-1em}
\end{wrapfigure}

\textbf{Grounding complexity.} Figure~\ref{fig:single_multi} compares overall mIoU ($\mathcal{M}$) scores on samples requiring single- vs. multi-instance masks for representative models (full results in Appendix~\ref{app:svm}). Most models degrade under multi-instance cases, highlighting simultaneous multi-region segmentation as a critical bottleneck. UniPixel and Sa2VA exhibit the most severe drops, despite achieving the strongest single-instance scores among grounding models. VRT-RL degrades least among grounding models ($\Delta \approx -7$). Among hybrid pipelines, Gemini(G)3+SAM3 declines sharply ($\Delta \approx -11$), whereas GPT-5.4+SAM3 maintains near-parity and slightly improves under multi-instance cases ($\Delta \approx +1$). Qwen3-VL-32B+SAM3 also improves on multi-instance samples ($\Delta \approx +4$), while Qwen3-VL-4B-I+SAM3 drops moderately, suggesting that model scale may help SAM3 for compositional grounding difficulty.

\section{Conclusion, Limitations, and Societal Impacts} \label{conclusion}

\textbf{Conclusion.}
We have introduced \ours, a benchmark for the joint evaluation of free-form  answer correctness and pixel-level evidence grounding in visual question answering. \ours spans six task types and six visual domains with explicit hallucination-aware scenarios. To support rigorous evaluation, we have proposed \metric, a unified metric that jointly assesses answer correctness and grounding fidelity via the geometric mean of smoothed per-axis scores. Comprehensive experiments across reasoning segmentation models, hybrid pipelines, and general-purpose MLLMs reveal that all evaluated models achieve low \metric scores, exposing a fundamental gap between answer fluency and visual grounding fidelity that existing benchmarks fail to capture. We hope \ours and \metric serve as a foundation for future research in grounded multimodal evaluation.

\textbf{Limitations.}
Due to the static nature of \ours, video and multi-image grounding scenarios are not covered. Generating high-quality samples that simultaneously satisfy answer correctness, mask fidelity, and task diversity is inherently challenging and requires rigorous manual control at every stage of the pipeline. As a result, \ours is limited to 1,157 high-quality samples despite significant annotation effort. This limitation stems from a circular dependency: the very models \ours is designed to evaluate are not yet reliable enough to automate its construction.

\textbf{Societal Impacts.} \ours is designed to improve the evaluation rigor of vision-language models, with direct benefits for transparency and reliability in safety-critical applications such as autonomous driving and robotics. The benchmark is constructed using images from publicly available datasets and contains no personally identifiable information. We have taken steps to mitigate annotation bias through multi-round human verification and cross-validation by independent annotators.

\section{Acknowledgements}
This work was supported by the Natural Sciences and Engineering Research Council of Canada (NSERC)’s Postdoctoral Fellowship, Mitacs Globalink Research Award, NSERC-CSE Research Community project entitled “An End-to-End Approach to Safe and Secure
AI Systems” and  NASA University Leadership Initiative. Any opinions, findings, and conclusions or recommendations expressed in this material are those of the authors and do not necessarily reflect the views of
any aforementioned supporting entity.

\small
\bibliographystyle{plainnat}
\bibliography{references}

\newpage
\appendix
\section{Appendix}

\begin{itemize} 

    \item [\ref{app:task}:] Tasks
    \item [\ref{app:ver}:] Quality Control
     \item [\ref{app:result}:] Additional Results
    \item [\ref{app:aggreg}:] Choice of Aggregation Function
    \item [\ref{app:epsilon}:] Choice of $\epsilon$
    \item [\ref{app:svm}:] Grounding Complexity
    \item [\ref{app:RSA}:] Ranking Stability Analysis
    \item [\ref{app:qual}:] Additional Qualitative Results
    \item [\ref{app:failure}:] Examples of Failure Cases in VQA Generation
    \item [\ref{app:data-construct}:] Prompts for Generating VQA Tasks 
    \item [\ref{app:judge}:] Prompt for LLM-as-a-Judge Evaluation of Answer Correctness 
    \item [\ref{app:output-format}:] Structured Output Format

\end{itemize}

\subsection{Tasks} \label{app:task}
\ours defines six task types that reflect the diverse 
reasoning demands of VQA, ranging from direct perception to multi-hop reasoning. Each task requires models to jointly produce a correct free-form answer and a segmentation mask that explicitly grounds the prediction in the relevant image 
region. The tasks include \emph{Identification}, 
\emph{Attribute}, \emph{OCR}, \emph{Spatial}, 
\emph{Counting}, and \emph{Comparison}, each 
described below.

\begin{enumerate}
    \item \textbf{Identification} evaluates the ability to recognize and name objects, entities, or scene elements present in the image. The task 
    requires precise localization of the target and accurate category-level or instance-level recognition, particularly in cluttered or multi-object scenes where discriminative grounding is essential.

    \item \textbf{Attribute} evaluates the ability to identify and describe specific properties of objects or regions, including color, shape, texture, and fine-grained appearance characteristics. Success requires attention to subtle visual details and accurate association of properties with the correct image region, particularly when multiple objects share similar features.

    \item \textbf{OCR} evaluates the ability to detect, read, and interpret text present in the scene, requiring tight integration of localization and 
    text recognition. This task is particularly challenging in domains where text appears at varying scales, orientations, or under partial occlusion, and where the answer must be grounded in the specific image region containing the relevant text. 
    
    \item \textbf{Spatial} evaluates the ability to interpret positional and geometric relationships between objects or regions in the scene, including absolute and relative positions, directional relationships, and proximity. The task requires integrating localization with relational reasoning to produce answers grounded in the correct spatial configuration.
    
    \item \textbf{Counting} evaluates the ability to enumerate instances of a specified category within the image, requiring systematic localization of 
    all relevant regions and aggregation of evidence across the scene. This task is particularly demanding in dense or occluded scenes where individual instances are difficult to discriminate.

    \item \textbf{Comparison} evaluates the ability to reason about differences or similarities between two or more objects, regions, or attributes within 
    the image. Correct answers require localizing relevant region, and performing relational inference over the evidence.
\end{enumerate}

\subsection{Quality Control}
\label{app:ver}
The annotation and verification process involved 12 human experts with diverse backgrounds and experience levels. All verification was conducted as part of standard research activities or voluntarily.

\textbf{Question–answer verification.} Annotators reviewed candidate question–answer pairs and selected the most semantically precise candidates based on four criteria: (1) the question is answerable using only the visual information in the image; (2) it is consistent with the intended task type, domain, and hallucination category; (3) the answer is factually correct with respect to the image; and (4) both the question and answer are clear and unambiguous.

\textbf{Mask verification.} Verifiers inspected segmentation masks to ensure that (1) the annotated regions correspond to the intended objects; (2) all relevant regions are included without extraneous areas; (3) mask boundaries accurately follow the target objects; and (4) hallucination-aware samples have empty ground-truth masks.

\begin{table*}[t]
\caption{Detailed per-task results on \ours. We report overall text accuracy ($\mathcal{T}$) and overall mIoU ($\mathcal{M}$) per task. $^\dagger$ \texttt{R-Sa2VA-Qwen3VL-4B-RL} checkpoint.}
\centering
\resizebox{\textwidth}{!}{
\begin{tabular}{l | cc cc cc cc cc cc}
\toprule
& \multicolumn{2}{c}{\textbf{Identification}} 
& \multicolumn{2}{c}{\textbf{Attribute}} 
& \multicolumn{2}{c}{\textbf{OCR}} 
& \multicolumn{2}{c}{\textbf{Spatial}} 
& \multicolumn{2}{c}{\textbf{Counting}} 
& \multicolumn{2}{c}{\textbf{Comparison}} \\
\cmidrule(lr){2-3}\cmidrule(lr){4-5}\cmidrule(lr){6-7}
\cmidrule(lr){8-9}\cmidrule(lr){10-11}\cmidrule(lr){12-13}
\textbf{Model} 
& $\mathcal{T}$ & $\mathcal{M}$
& $\mathcal{T}$ & $\mathcal{M}$
& $\mathcal{T}$ & $\mathcal{M}$
& $\mathcal{T}$ & $\mathcal{M}$
& $\mathcal{T}$ & $\mathcal{M}$
& $\mathcal{T}$ & $\mathcal{M}$ \\
\midrule
LISA-7B~\cite{lai2024lisa}   & 7.48 &  16.37 & 12.04 & 19.85 & 2.89 & 9.24 & 5.03 & 19.34 & 8.51 & 22.05 & 7.29 & 19.36 \\					
SESAME-7B~\cite{wu2024sesame} & 3.74 & 26.80 & 2.62 & 26.32 & 2.31 & 12.98 & 5.03 & 23.57 & 3.19 & 25.59 & 2.08 & 25.40 \\ 					
GLaMM-7B~\cite{rasheed2024glamm} & 2.34 & 19.72 & 1.57 & 11.63 & 0.58 & 10.51 & 1.01 & 17.27 & 0.53 & 9.27 & 2.08 & 18.51 \\					
LaSagnA-7B~\cite{wei2024lasagna}  & 0.00 & 15.41 & 0.00 & 9.80 & 0.58 & 4.96 & 1.01 & 13.82 & 0.53 & 6.03 & 0.52 & 13.10 \\ 					
UniPixel-7B~\cite{liuunipixel}   & 17.76 & 24.77 & 30.89 & 25.05 & 32.95 & 13.72 & 11.56 & 23.21 & 10.64 & 9.87 & 25.52 & 24.53 \\
Sa2VA-8B~\cite{yuan2025sa2va} & 27.10 & 26.29 & 33.51 & 26.94 & 41.62 & 12.11 & 17.59 & 22.86 & 25.53 & 11.84 & 30.73 & 24.32 \\ 					
VRT-RL$^\dagger$~\cite{yuan2025vrt} & 39.25 &  30.32 & 39.79 & 23.22 & 38.15 & 24.75 & 28.64 & 19.82 & 28.72 & 27.94 & 43.23 & 19.06 \\				
TreeVGR-7B~\cite{wang2026treebench} & 11.21 & 23.15 & 14.66 & 20.46 & 9.25 & 18.75 & 12.56 & 17.79 & 18.62 & 19.43 & 30.21 & 17.96 \\ 					
Uground-7B~\cite{qian2025uground}      & 0.47 & 4.70 & 1.05 & 5.99 & 1.16 & 4.45 & 0.50 & 6.74 & 0.53 & 10.38 & 2.08 & 11.03 \\					
\midrule
Qwen3-VL-4B-I + SAM3    & 54.67 & 48.53 & 58.12 & 43.89 & 63.01 & 31.47 & 41.21 & 37.95 & 46.28 & 47.13 & 56.77 & 28.92 \\	
Qwen3-VL-32B-I + SAM3  & 58.88 & 43.61 & 60.73 & 42.10 & 68.79 & 29.70 & 48.74 & 27.43 & 47.34 & 47.83 & 62.50 & 24.54 \\
Gemini 3  + SAM3   & 63.55 & 39.56 & 60.21 & 39.51 & 72.83 & 33.54 & 49.75 & 34.01 & 61.70 & 32.19 & 67.19 & 30.48 \\
GPT-5.4 + SAM3    & 57.48 & 47.98 & 49.21 & 42.52 & 65.32 & 32.54 & 47.24 & 30.84 & 46.28 & 53.98 & 56.25 & 26.67 \\
GPT-5.4-T + SAM3 & 63.08 & 54.65 & 59.16 & 43.55 & 73.41 & 33.20 & 51.26 & 38.59 & 55.85 & 51.85 & 64.58 & 28.09 \\

\bottomrule
\end{tabular}
}
\label{app:tab_task_results}
\vspace{-1.5pt}
\end{table*}

\begin{table*}[t]
\caption{Detailed per-domain results on \ours. We report overall text accuracy ($\mathcal{T}$) and overall mIoU ($\mathcal{M}$) per domain.$^\dagger$ \texttt{R-Sa2VA-Qwen3VL-4B-RL} checkpoint.}
\centering
\resizebox{\textwidth}{!}{
\begin{tabular}{l | cc cc cc cc cc cc}
\toprule
& \multicolumn{2}{c}{\textbf{Indoor}} 
& \multicolumn{2}{c}{\textbf{Outdoor}} 
& \multicolumn{2}{c}{\textbf{Math}} 
& \multicolumn{2}{c}{\textbf{AV}} 
& \multicolumn{2}{c}{\textbf{Robotics}} 
& \multicolumn{2}{c}{\textbf{Science}} \\
\cmidrule(lr){2-3}\cmidrule(lr){4-5}\cmidrule(lr){6-7}
\cmidrule(lr){8-9}\cmidrule(lr){10-11}\cmidrule(lr){12-13}
\textbf{Model} 
& $\mathcal{T}$ & $\mathcal{M}$
& $\mathcal{T}$ & $\mathcal{M}$
& $\mathcal{T}$ & $\mathcal{M}$
& $\mathcal{T}$ & $\mathcal{M}$
& $\mathcal{T}$ & $\mathcal{M}$
& $\mathcal{T}$ & $\mathcal{M}$ \\
\midrule
LISA-7B~\cite{lai2024lisa}   & 	6.34 & 19.41 & 11.22 & 18.56 & 2.67 & 8.74 & 11.88 & 18.18 & 6.00 & 21.32 & 4.41 & 18.33 \\ 									
SESAME-7B~\cite{wu2024sesame}  &  3.90 & 22.40 & 4.59 & 23.41 & 0.67 & 4.58 & 4.95 & 35.24 & 2.00 & 22.18 & 2.45 & 29.23 \\										
GLaMM-7B~\cite{rasheed2024glamm} & 2.44 & 9.12 & 2.04 & 15.35 & 0.00 & 6.50 & 1.98 & 23.83 & 1.00 & 14.53 & 0.49 & 16.78 \\										
LaSagnA-7B~\cite{wei2024lasagna}  & 0.00 & 6.77 & 1.53 & 8.87 & 0.67 & 5.99 & 0.50 & 17.40 & 0.00 & 9.90 & 0.00 & 14.24 \\ 								
UniPixel-7B~\cite{liuunipixel}  & 26.34 & 15.43 & 23.47 & 19.66 & 3.33 & 11.46 & 29.21 & 29.56 & 14.50 & 21.56 & 25.98 & 22.65 \\
Sa2VA-8B~\cite{yuan2025sa2va}  &  29.27 & 15.59 & 31.63 & 17.60 & 10.00 & 9.65 & 44.55 & 32.70 & 26.00 & 21.91 & 27.94 & 25.63 \\										
VRT-RL$^\dagger$~\cite{yuan2025vrt}  & 40.98 & 21.69 & 39.80 & 27.89 & 28.00 & 18.75 & 41.09 & 27.30 & 29.50 & 22.88 & 36.27 & 25.71 \\ 										
TreeVGR-7B~\cite{wang2026treebench}  & 16.59 & 16.51 & 23.47 & 20.13 & 4.67 & 9.15 & 21.78 & 27.83 & 14.00 & 24.70 & 13.24 & 17.08 \\  								
Uground-7B~\cite{qian2025uground}   & 0.49 & 5.92 & 0.51 & 3.45 & 0.00 & 4.80 & 0.99 & 7.41 & 1.50 & 12.31 & 1.96 & 8.64 \\ 										
\midrule
Qwen3-VL-4B-I + SAM3   &  52.20 & 39.45 & 58.67 & 42.21 & 35.33 & 22.50 & 72.28 & 55.33 & 45.00 & 43.01 & 50.98 & 32.67 \\	
Qwen3-VL-32B-I + SAM3   & 61.95 & 34.83 & 65.31 & 39.47 & 32.67 & 17.18 & 76.73 & 49.55 & 44.00 & 41.26 & 58.82 & 29.22 \\
Gemini 3 + SAM3   & 62.93 & 34.79 & 63.27 & 33.24 & 55.33 & 22.97 & 67.33 & 46.48 & 51.50 & 37.20 & 71.57 & 32.22 \\
GPT-5.4 + SAM3  & 62.93 & 39.00 & 63.27 & 39.74 & 55.33 & 23.30 & 67.33 & 51.67 & 51.50 & 45.37 & 71.57 & 32.52 \\
GPT-5.4-T + SAM3  & 64.88 & 39.58 & 65.31 & 41.55 & 58.00 & 33.37 & 68.81 & 48.89 & 48.00 & 50.15 & 60.29 & 36.31 \\

\bottomrule
\end{tabular}
}
\label{app:tab_domain_results}
\vspace{-10 pt}
\end{table*}

\textbf{Cross-verification and disagreement resolution.} Each question–answer–mask triplet underwent an additional review by a second independent verifier. Ambiguous cases or disagreements were escalated to at least two senior researchers for discussion, with the final decision determined by consensus.

\subsection{Additional Results} \label{app:result}
Table~\ref{app:tab_task_results} reports overall text and mask scores per task type. While pipeline models achieve strong text scores, particularly on OCR tasks (\eg, Gemini 3 + SAM3 reaches 72.83), their corresponding mask scores remain low. Crucially, high text and mask scores do not necessarily reflect correct performance on the same samples: a model may answer the textual query correctly on one subset of samples while producing accurate masks on a entirely different subset. This decoupling is precisely why \metric, reported in the main paper, is needed as a joint metric that rewards models only when both modalities are correct for the same instance.

Table~\ref{app:tab_domain_results} reveals the same pattern across all six domains, where dedicated grounding and segmentation models consistently underperform hybrid pipeline approaches. AV scenes yield the highest text scores for pipeline models, yet mask scores there remain moderate, while Math is the hardest domain across both modalities for all model families. Crucially, the persistent text-mask gap across all domains reinforces that reporting either score in isolation paints an incomplete and potentially misleading picture of model capability, further validating GROVE as the primary evaluation metric.

\subsection{Choice of Aggregation Function} \label{app:aggreg}
To assess the robustness of \metric to the choice of aggregation function, we recompute per-sample joint scores for all 14 evaluated models using the geometric mean, harmonic mean, and raw product, applying the same $\epsilon$-flooring. As shown in Table~\ref{tab:agg_sens}, all three formulations yield identical model rankings (Spearman's $\rho=1.0$, Kendall's $\tau=1.0$), despite differences in absolute scores. This confirms that our comparative conclusions are not dependent on the specific aggregation function, while the geometric mean provides a balance between penalizing imbalanced predictions and maintaining an interpretable score range.

\begin{table*}[t]
\centering
\caption{Sensitivity to aggregation functions. All three
formulations use the same $\epsilon$ and yield
identical model rankings (Spearman's $\rho=1.0$,
Kendall's $\tau=1.0$).}
\label{tab:agg_sens}
\begin{tabular}{l|ccc}
\toprule
\textbf{Model} & \textbf{GROVE}
& \textbf{Harmonic Mean} & \textbf{Product} \\
\midrule
GPT-5.4-T + SAM3 & 45.53 & 39.42 & 30.99 \\
Qwen3-VL-4B-I + SAM3 & 43.30 & 38.53 & 30.64 \\
Qwen3-VL-32B-I + SAM3 & 42.72 & 37.54 & 29.29 \\
GPT-5.4 + SAM3 & 41.50 & 35.77 & 27.03 \\
Gemini 3 + SAM3 & 39.63 & 31.53 & 23.93 \\
VRT-RL & 29.02 & 24.44 & 15.40 \\
Sa2VA & 25.74 & 22.14 & 12.87 \\
UniPixel & 22.39 & 19.10 & 9.72 \\
TreeVGR-7B & 19.03 & 15.33 & 5.70 \\
SESAME & 17.02 & 14.35 & 5.54 \\
LISA-7B & 15.47 & 12.34 & 3.29 \\
GLaMM & 13.53 & 11.73 & 2.39 \\
LaSagnA & 12.35 & 11.15 & 1.87 \\
UGround-7B & 11.78 & 10.94 & 1.84 \\
\bottomrule
\end{tabular}
\vspace{-5pt}
\end{table*}

\subsection{Choice of \texorpdfstring{$\epsilon$}{epsilon}} \label{app:epsilon}
The flooring parameter $\epsilon$ in \metric sets the minimum score assigned to failed predictions, preventing score collapse during aggregation. More specifically, this parameter prevents the joint score from collapsing to zero when one component fails, preserving the ability to discriminate models based on the quality of the other component. We perform a sensitivity analysis on $\epsilon$ with results shown in Tables~\ref{app:tab_sensitivity_scores} and~\ref{app:tab_sensitivity_rho} for  $\epsilon \in \{0.01, 0.05, 0.1\}$. We see that \metric exhibits strong robustness to the choice of the flooring parameter $\epsilon$. While absolute \metric scores increase monotonically with larger $\epsilon$, model rankings are fully preserved across all settings ($\rho = 1.00$), suggesting that the relative ordering of models is invariant for $\epsilon \in [0.01, 0.1]$. 

This stability is further supported by the correlation analysis in Table~\ref{app:tab_sensitivity_rho}. At finer granularity, per-task and per-domain correlations remain high ($\rho \geq 0.971$). Accordingly, we adopt $\epsilon = 0.1$, which provides the highest score discriminability while preserving ranking consistency.

While larger values of $\epsilon$ monotonically increase absolute 
\metric scores and preserve model rankings (as shown in Table~\ref{app:tab_sensitivity_scores}), 
setting $\epsilon > 0.1$ progressively undermines the metric's core desideratum of \textit{joint sensitivity}. To see why, consider the extreme case $\epsilon \to 1$: both $S'_a$ and $S'_m$ collapse toward $1$ regardless of actual model performance, rendering \metric uninformative. More concretely, at $\epsilon = 0.1$ a model that fails entirely on one modality (\eg, $S_a = 0$) receives a floored score of $S'_a = 0.1$, contributing a joint score of at most $\sqrt{0.1 \times 1} \approx 0.316$ — a meaningful penalty. At $\epsilon = 0.3$, the same fully-failing model on one modality would receive $\sqrt{0.3 \times 1} \approx 0.548$, which reduces the headroom to distinguish it from models with genuine partial competence. We therefore choose $\epsilon = 0.1$ 
as the largest value that preserves meaningful penalization of 
single-modality failures while avoiding score collapse in the 
other direction.

The use of $\epsilon$-flooring ensures that samples with one failed component do not collapse to zero, while still penalizing the lack of joint correctness. This can result in seemingly anomalous results, where per-sample scores \eqref{eq:per-sample} exceed one or both of the individual components and the \metric score \eqref{eq:grove} is not bounded by the dataset-level averages of its marginals. This is intentional: \metric deliberately reflects a softened measure of joint alignment.

\begin{table*}[t]
\caption{\metric scores and ranks for all models across 
$\epsilon \in \{0.01, 0.05, 0.1\}$. Rankings are fully 
preserved across all settings ($\rho = 1.00$), with 
$\epsilon = 0.1$ yielding the highest discriminability. $^\dagger$ \texttt{R-Sa2VA-Qwen3VL-4B-RL} checkpoint.}
\centering
\begin{tabular}{l | c c c | c c c}
\toprule
& \multicolumn{3}{c|}{\textbf{\metric Score}} 
& \multicolumn{3}{c}{\textbf{Rank}} \\
\cmidrule{2-4} \cmidrule{5-7}
\textbf{Model} 
& $\epsilon=0.01$ & $\epsilon=0.05$ & $\epsilon=0.1$ 
& $\epsilon=0.01$ & $\epsilon=0.05$ & $\epsilon=0.1$ \\
\midrule
LISA-7B~\cite{lai2024lisa}              &  3.61 &  9.60 & 15.47 & 11 & 11 & 11 \\
SESAME-7B~\cite{wu2024sesame}           &  5.87 & 11.44 & 17.02 & 10 & 10 & 10 \\
GLaMM-7B~\cite{rasheed2024glamm}        &  2.93 &  8.15 & 13.53 & 12 & 12 & 12 \\
LaSagnA-7B~\cite{wei2024lasagna}        &  2.37 &  7.16 & 12.35 & 13 & 13 & 13 \\
UniPixel-7B~\cite{liuunipixel}          & 12.02 & 17.24 & 22.39 &  8 &  8 &  8 \\
Sa2VA-8B~\cite{yuan2025sa2va}           & 16.02 & 20.88 & 25.74 &  7 &  7 &  7 \\
VRT-RL$^\dagger$~\cite{yuan2025vrt}     & 17.93 & 23.70 & 29.02 &  6 &  6 &  6 \\
TreeVGR-7B~\cite{wang2026treebench}     &  7.54 & 13.51 & 19.03 &  9 &  9 &  9 \\
UGround-7B~\cite{qian2025uground}       &  2.13 &  6.64 & 11.78 & 14 & 14 & 14 \\
\midrule
Qwen3-VL-4B + SAM3       & 34.14 & 38.98 & 43.30 &  2 &  2 &  2 \\
Qwen3-VL-32B + SAM3      & 33.51 & 38.40 & 42.72 &  3 &  3 &  3 \\
Gemini-3-Flash + SAM3    & 25.49 & 33.30 & 39.63 &  5 &  5 &  5 \\
GPT-no-thinking + SAM3   & 31.61 & 36.90 & 41.50 &  4 &  4 &  4 \\
GPT-thinking + SAM3      & 35.86 & 41.08 & 45.53 &  1 &  1 &  1 \\
\bottomrule
\end{tabular}
\label{app:tab_sensitivity_scores}
\vspace{-5 pt}
\end{table*}

\begin{table*}[t]
\caption{Spearman $\rho$ between \metric rankings across
$\epsilon$ values at overall, per-task, and per-domain
granularity. All correlations are statistically significant.}
\centering
\begin{tabular}{l l | c c c}
\toprule
& \textbf{Granularity}
& $\boldsymbol{\epsilon_{0.1}}$ vs $\boldsymbol{\epsilon_{0.05}}$
& $\boldsymbol{\epsilon_{0.1}}$ vs $\boldsymbol{\epsilon_{0.01}}$
& $\boldsymbol{\epsilon_{0.05}}$ vs $\boldsymbol{\epsilon_{0.01}}$ \\
\midrule
& Overall        & 1.000 & 1.000 & 1.000 \\
\midrule
\multirow{6}{*}{\rotatebox[origin=c]{90}{\textbf{Task}}}
& Identification & 1.000 & 0.996 & 0.996 \\
& Attribute      & 0.996 & 0.993 & 0.996 \\
& OCR            & 0.993 & 0.971 & 0.986 \\
& Spatial        & 1.000 & 0.996 & 0.996 \\
& Counting       & 0.989 & 0.986 & 0.996 \\
& Comparison     & 0.996 & 0.986 & 0.993 \\
\midrule
\multirow{6}{*}{\rotatebox[origin=c]{90}{\textbf{Domain}}}
& Indoor         & 0.996 & 0.996 & 1.000 \\
& Outdoor        & 1.000 & 1.000 & 1.000 \\
& Math           & 0.988 & 0.974 & 0.982 \\
& AV             & 0.996 & 0.993 & 0.996 \\
& Robotics       & 0.993 & 0.993 & 1.000 \\
& Science        & 0.996 & 0.968 & 0.975 \\
\bottomrule
\end{tabular}
\label{app:tab_sensitivity_rho}
\vspace{-5 pt}
\end{table*}

\subsection{Grounding Complexity} \label{app:svm}

\begin{figure*}[ht]
\centering
\includegraphics[width=0.75\textwidth]{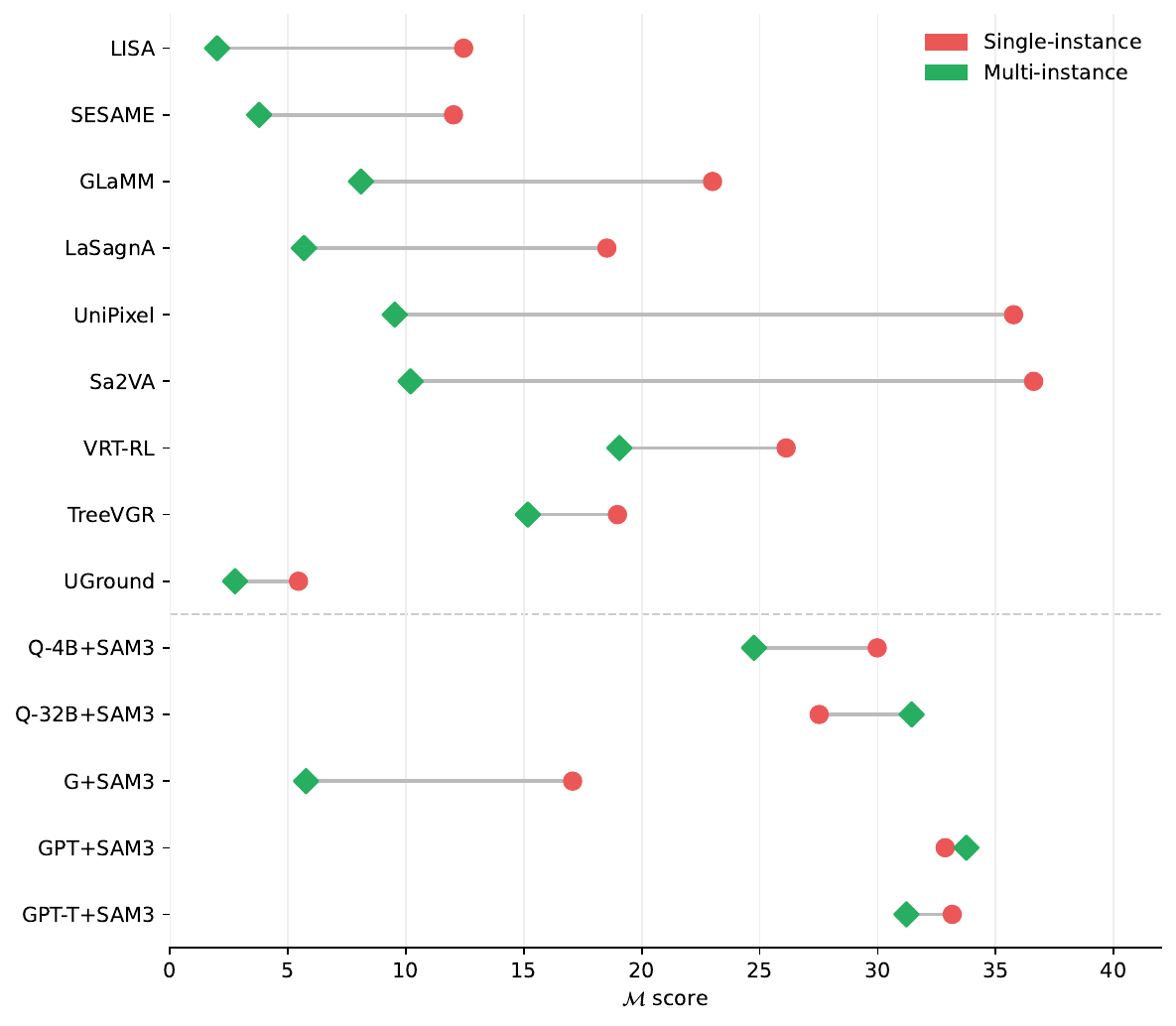}
\caption{Single- vs.\ multi-instance overall mask 
scores for all 14 models. The dashed line separates grounding-specific models (above) 
from hybrid pipelines (below).}
\label{app:fig_svm}
\vspace{-15 pt}
\end{figure*}

Figure~\ref{app:fig_svm} extends the analysis 
in the main paper to all 14 models, reporting overall mask scores for single- and multi-instance samples 
separately. The degradation pattern is consistent across 
model families: grounding-specific models show the largest 
absolute drops, with UniPixel and Sa2VA collapsing by 
over 26 points, while TreeVGR and VRT-RL are comparatively 
robust. Among hybrid pipelines, Gemini3+SAM3 degrades 
sharply despite strong overall performance, whereas 
GPT-5.4+SAM3 and Qwen3-VL-32B+SAM3 improve 
their mask scores under multi-instance grounding, 
suggesting that model scale and reasoning ability 
partially compensate for the added segmentation complexity. 
These results confirm that multi-instance grounding 
represents a fundamentally harder challenge than 
single-instance localization across all evaluated 
model families.

\subsection{Ranking Stability Analysis.} \label{app:RSA}

To assess whether the size of \ours is sufficient for reliable model comparison, we conduct subsampling and bootstrap analyses of ranking stability. First, we compare the full-benchmark model ranking with rankings obtained from random subsamples of varying sizes ($n=100$--$1000$), using 1,000 draws without replacement per size. As shown in Table~\ref{tab:stability}, mean Kendall's $\tau$ increases from 0.930 at $n=100$ to 0.994 at $n=750$ and 0.999 at $n=1000$, suggesting that rankings largely stabilize before reaching the full benchmark size.

We further perform bootstrap analysis using 5,000 resamples with replacement. The resulting rankings remain highly stable for both the full benchmark (mean $\tau=0.990$, 95\% interval [0.956, 1.000]) and the hallucination-aware subset (mean $\tau=0.980$, 95\% interval [0.934, 1.000]). Across the six task types and six visual domains, mean Kendall's $\tau$ exceeds 0.91, with the lowest values observed for Comparison ($\tau=0.91$, $n=192$), OCR ($\tau=0.91$, $n=173$), and Math ($\tau=0.92$, $n=150$). The reported 95\% intervals correspond to the 2.5th and 97.5th percentiles across resamples.

Overall, these results support the stability of relative model rankings at the benchmark and reported subset levels, despite the moderate dataset size. However, finer-grained comparisons involving smaller subsets should be interpreted with greater caution.

\begin{table}[t]
\centering
\caption{Ranking stability. Kendall's $\tau$ compares rankings obtained from
subsamples with the full-benchmark ranking.
Results are based on 1,000 draws without replacement
per sample size.}
\label{tab:stability}
\begin{tabular}{ccc}
\toprule
\textbf{\# Samples} & \textbf{Mean $\tau$}
& \textbf{95\% Interval} \\
\midrule
100  & 0.930 & [0.846, 1.000] \\
250  & 0.967 & [0.912, 1.000] \\
500  & 0.986 & [0.956, 1.000] \\
750  & 0.994 & [0.978, 1.000] \\
1000 & 0.999 & [0.978, 1.000] \\
\bottomrule
\end{tabular}
\vspace{-10 pt}
\end{table}

\subsection{Additional Qualitative Results} \label{app:qual}

\begin{figure*}
    \centering
    \includegraphics[width=0.95\textwidth]{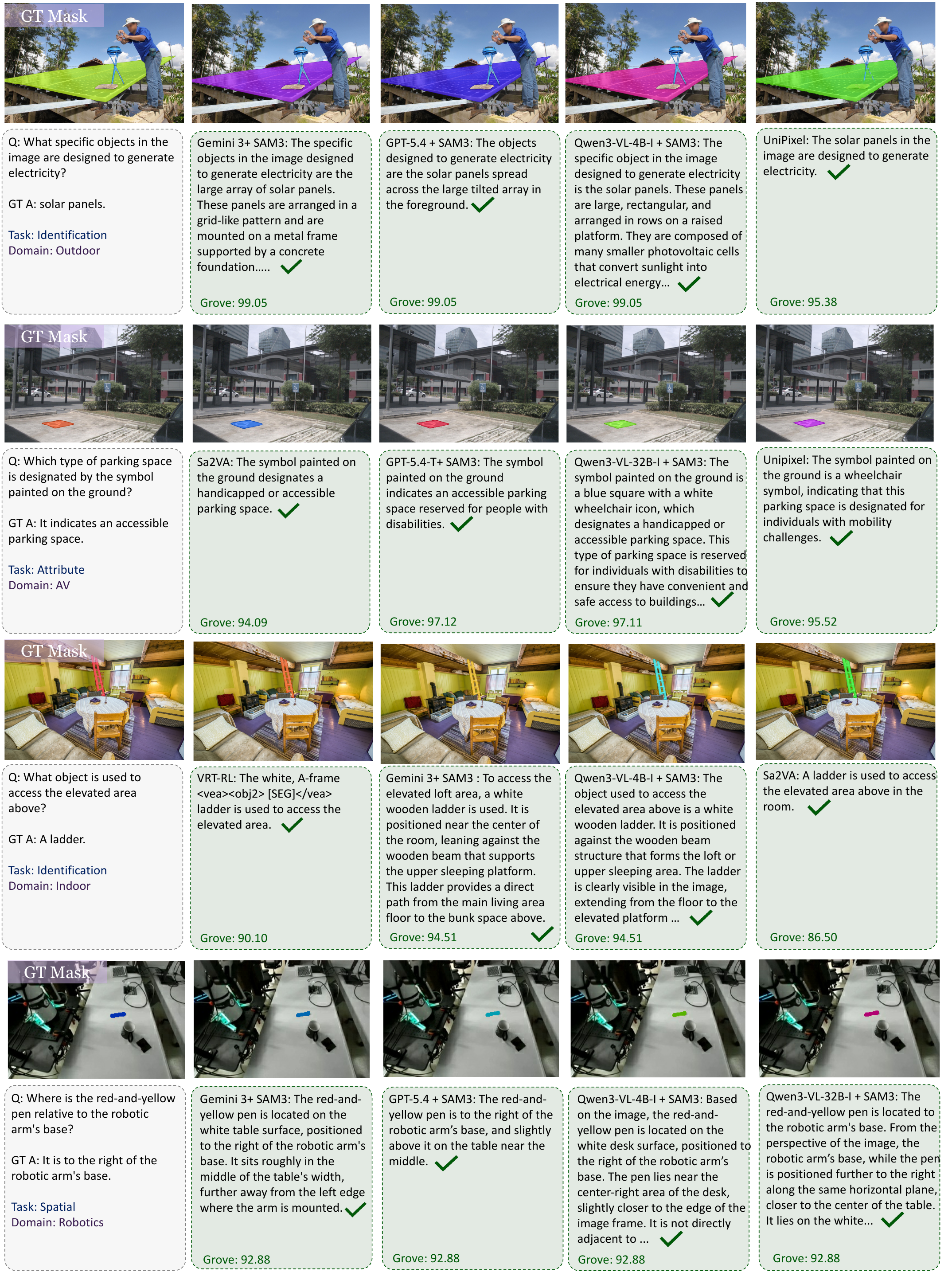}
    \vspace{-0.5em}
    \caption{
    Qualitative examples where the textual answer is correct (\textcolor{green!50!black}{$S_a = 1$}) and the evidence mask achieves \textcolor{green!50!black}{high} overlap with the ground-truth annotation, yielding high \metric scores.
    }
    \label{app:fig_bothc}
   \vspace{-1em}
\end{figure*}

\begin{figure*}
    \centering
    \includegraphics[width=0.95\textwidth]{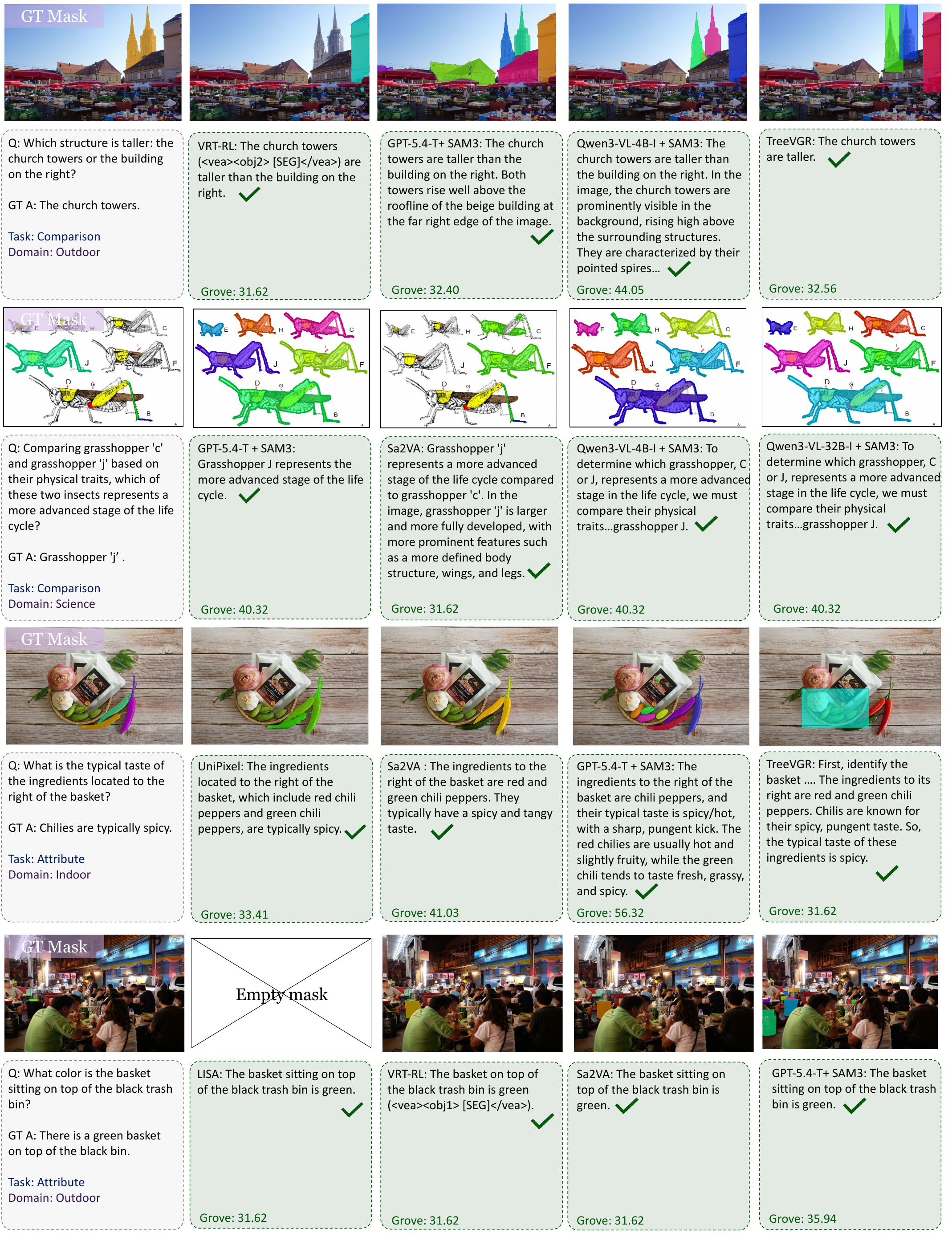}
    \vspace{-0.5em}
    \caption{
    Qualitative examples where the textual answer is correct (\textcolor{green!50!black}{$S_a = 1$}) and the evidence mask achieves \textcolor{red}{low} overlap with the ground-truth annotation, yielding low \metric scores.
    }
    \label{app:fig_t_high}
   \vspace{-1em}
\end{figure*}

\begin{figure*}
    \centering
    \includegraphics[width=0.95\textwidth]{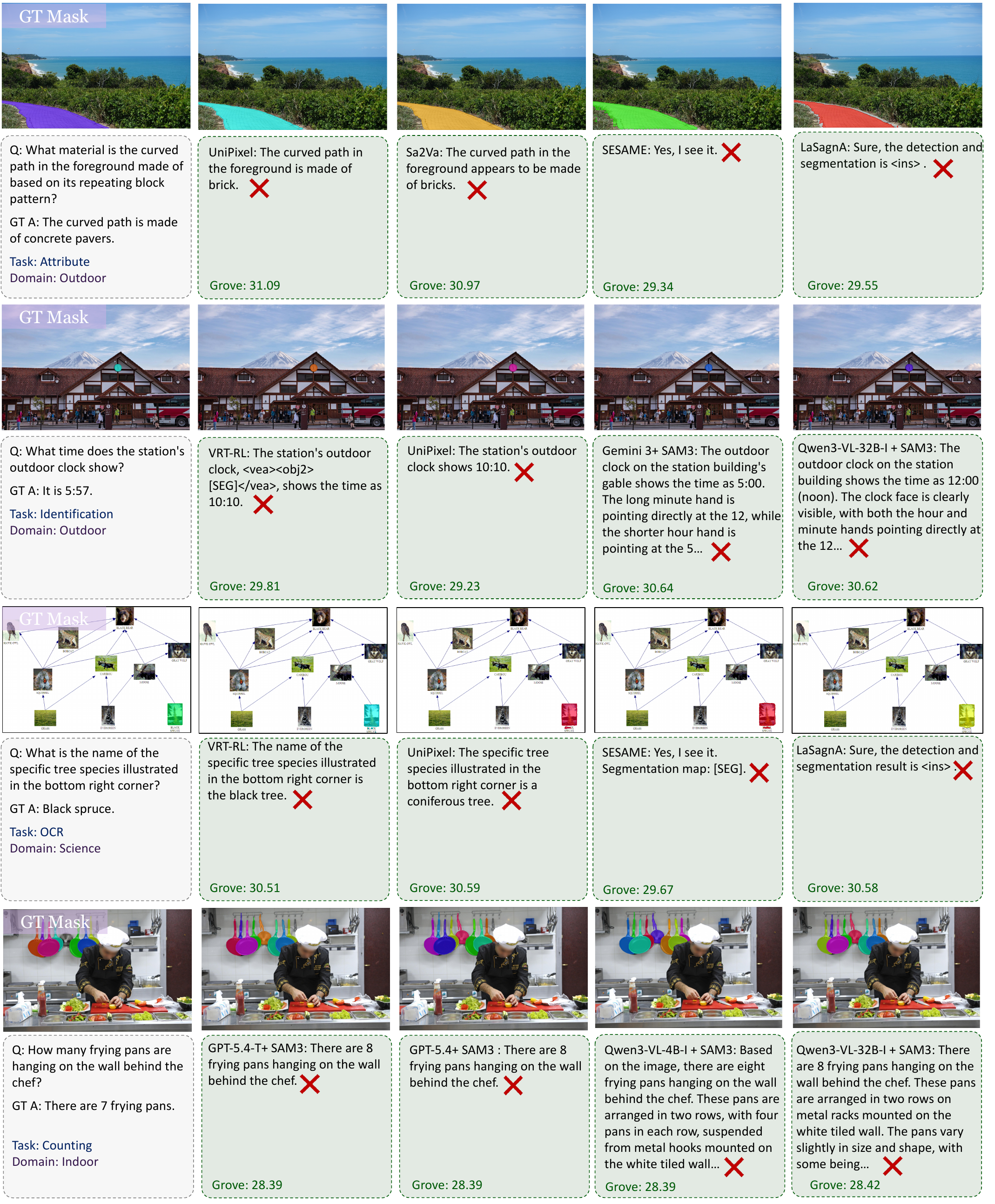}
    \vspace{-0.5em}
    \caption{
    Qualitative examples where the textual answer is incorrect (\textcolor{red}{$S_a = 0$}) and the evidence mask achieves \textcolor{green!50!black}{high} overlap with the ground-truth annotation, yielding low \metric scores.
    }
    \label{app:fig_m_high}
   \vspace{-1em}
\end{figure*}

\begin{figure*}
    \centering
    \includegraphics[width=0.95\textwidth]{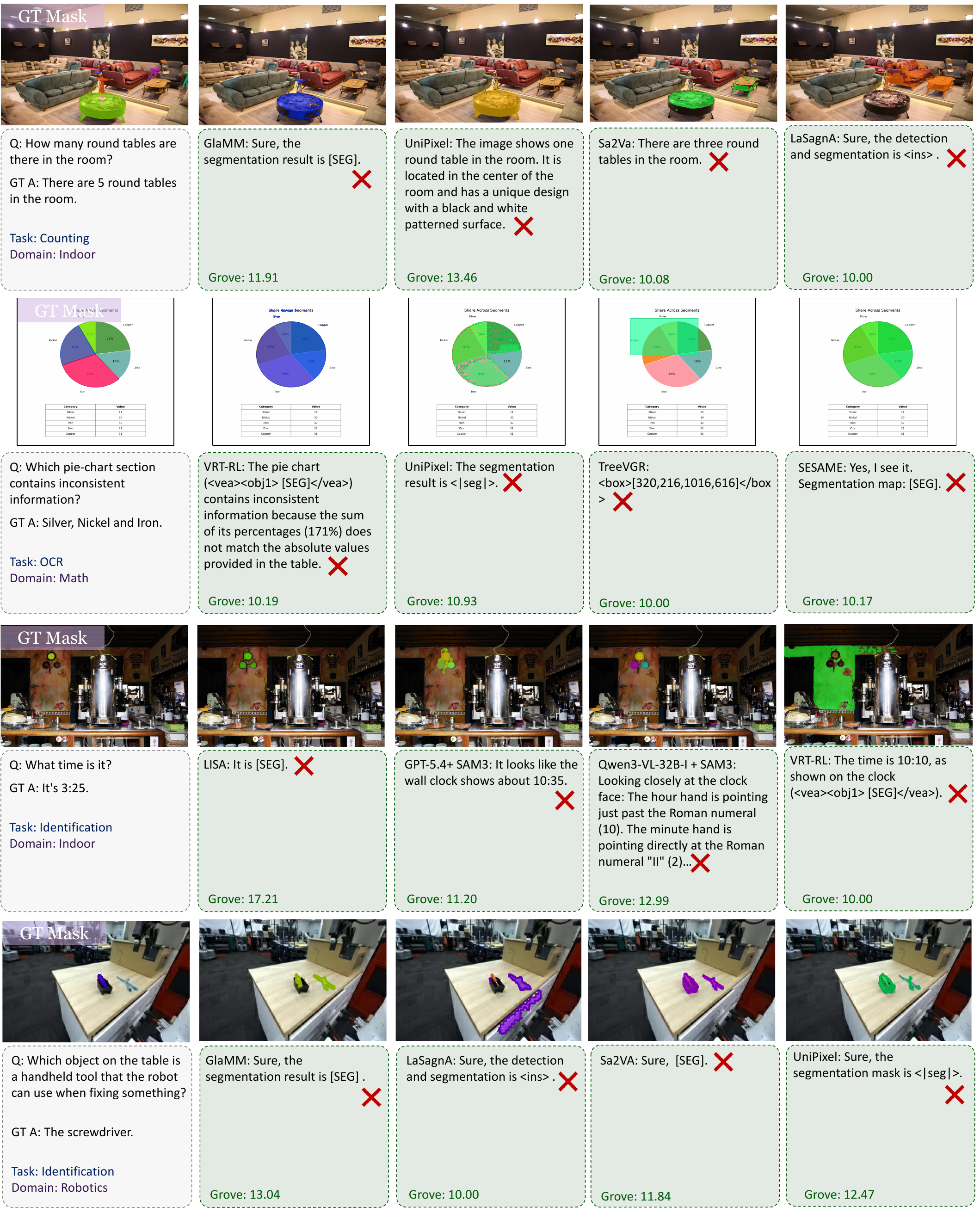}
    \vspace{-0.5em}
    \caption{
    Qualitative examples where the textual answer is incorrect (\textcolor{red}{$S_a = 0$}) and the evidence mask achieves \textcolor{red}{low} overlap with the ground-truth annotation, yielding low \metric scores.
    }
    \label{app:fig_bothi}
   \vspace{-1em}
\end{figure*}

\begin{figure*}
    \centering
    \includegraphics[width=0.95\textwidth]{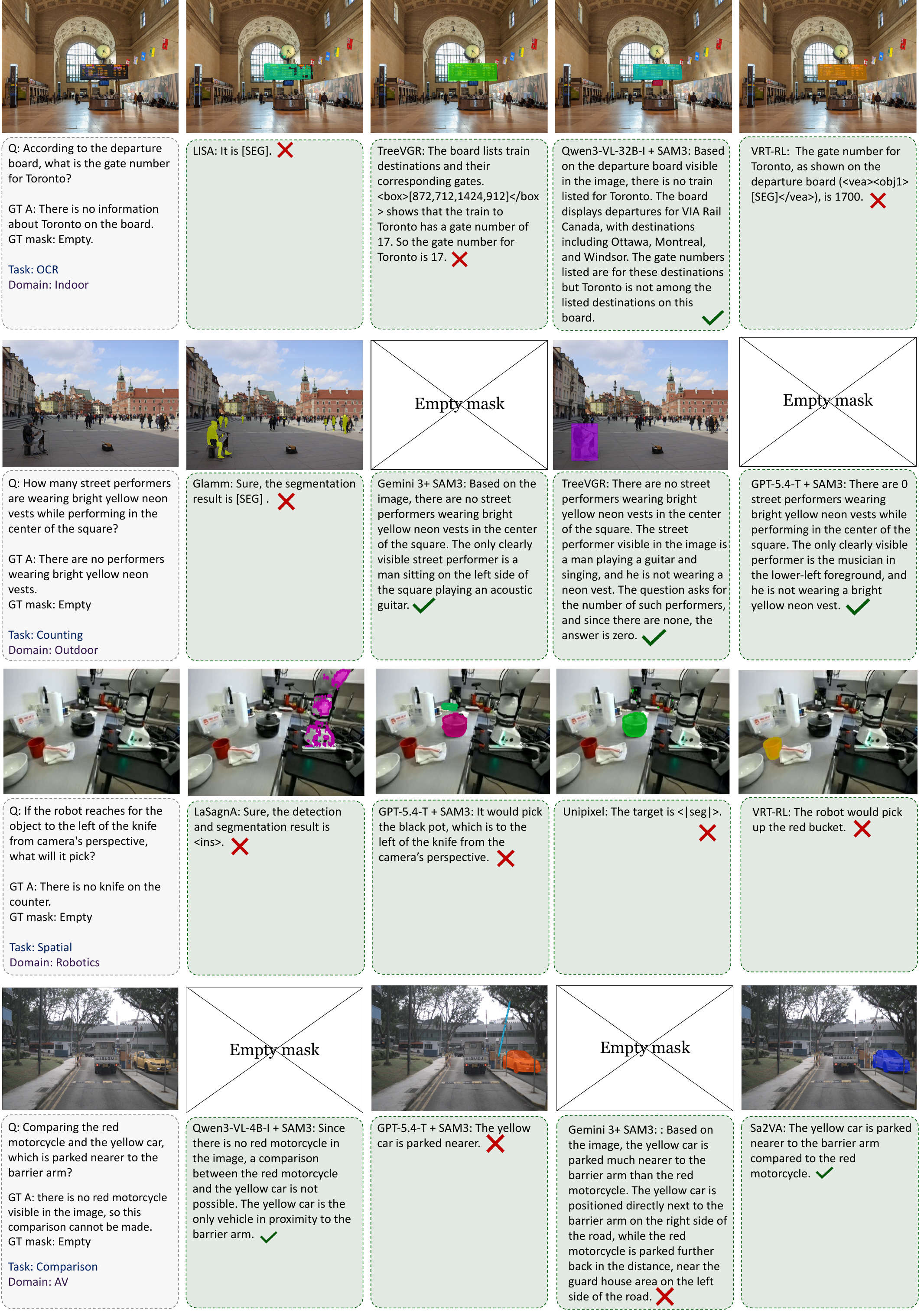}
    \vspace{-0.5em}
    \caption{
    Examples of various models' outputs on hallucination-aware samples, where the ground-truth mask is empty.
    }
    \label{app:fig_hallu}
   \vspace{-1em}
\end{figure*}

Figures~\ref{app:fig_bothc}--\ref{app:fig_hallu} present qualitative examples illustrating the four outcome combinations captured by \metric, along with hallucination-aware samples. Figure~\ref{app:fig_bothc} shows cases where both the textual answer and the evidence mask are correct, yielding high \metric scores (\eg, \metric $\approx 86$–$99$). These examples demonstrate that when models correctly answer the question and ground the target evidence precisely, \metric effectively rewards joint alignment across diverse tasks and domains, including outdoor scenes, robotics, and indoor environments.

Figure~\ref{app:fig_t_high} highlights a key failure mode: the textual answer is correct but the evidence mask is misaligned, resulting in relatively low \metric scores despite accurate reasoning. This reflects cases where models answer correctly but fail to localize the relevant evidence, confirming that \metric penalizes ungrounded correct answers. Conversely, Figure~\ref{app:fig_m_high} shows cases where the mask quality is high but the textual answer is incorrect, models segment the correct region but draw wrong conclusions, also yielding relatively low \metric scores.

Figure~\ref{app:fig_bothi} presents the worst-case scenario, where both the answer and the mask are incorrect, leading to the lowest \metric scores (\metric $\approx 10$–$13$). Notably, these scores do not collapse to zero due to $\epsilon$-flooring, as discussed in Section~\ref{sec:results}. Finally, Figure~\ref{app:fig_hallu} illustrates hallucination-aware samples where the ground-truth mask is empty. Models that correctly recognize the absence of the queried entity produce an empty mask, while those that hallucinate a non-empty mask or provide an incorrect answer are penalized.

Overall, these examples demonstrate that \metric captures a richer and more reliable signal than either text accuracy or mask quality alone.

\subsection{Examples of Failure Cases in VQA Generation} \label{app:failure}

\begin{figure*}[ht]
\centering
\begin{subfigure}[b]{0.85\textwidth}
    \centering
    \includegraphics[width=\textwidth]{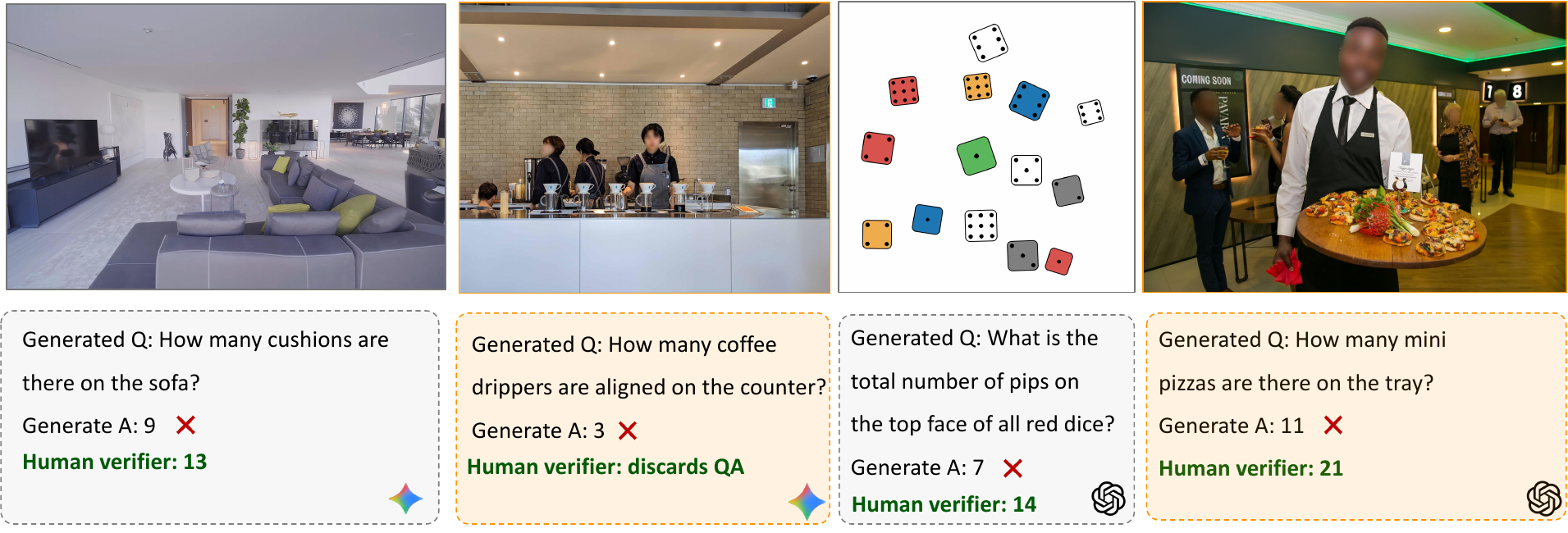}
    \caption{Counting-related failures in QA generation. }
    \label{app:fig_g_1}
\end{subfigure}

\vspace{1em}

\begin{subfigure}[b]{0.85\textwidth}
    \centering
    \includegraphics[width=\textwidth]{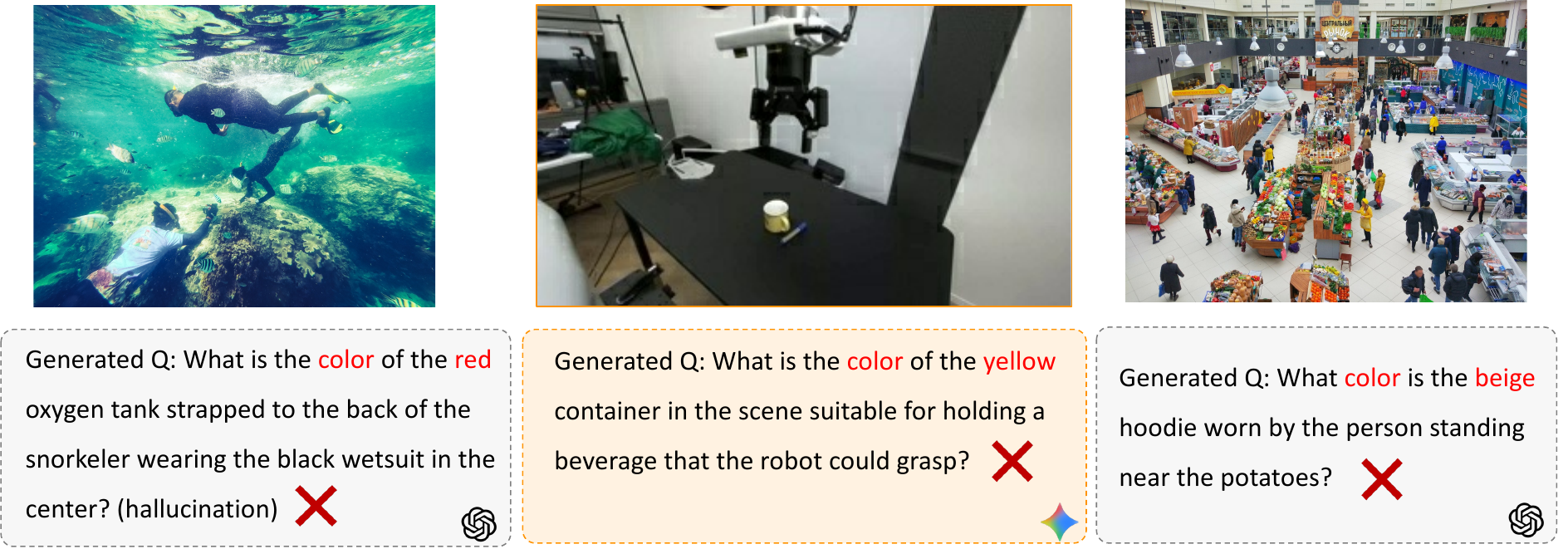}
    \caption{QA generation failures where the answer is implicitly contained in the question.}
    \label{app:fig_g_2}
\end{subfigure}

\vspace{1em}

\begin{subfigure}[b]{0.85\textwidth}
    \centering
    \includegraphics[width=\textwidth]{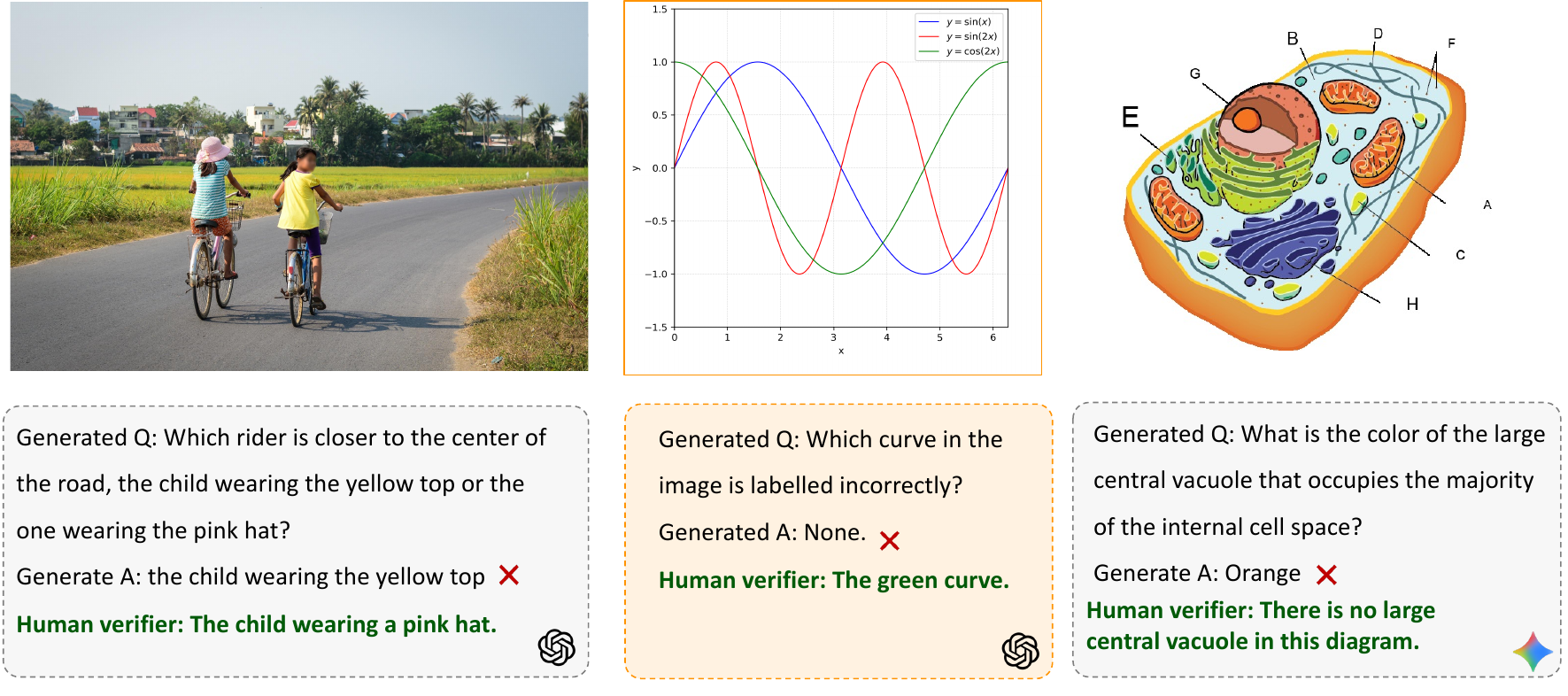}
    \caption{QA generation failures due to incorrect answers.}
    \label{app:fig_g_3}
\end{subfigure}

\caption{Failure cases in automated QA generation, highlighting the need for human verification}
\label{app:fig_g}
\end{figure*}

Figures~\ref{app:fig_g} present representative failure cases in QA generation using GPT-5.2~\cite{openai2025gpt52} and Gemini 3 pro~\cite{google2025gemini3}, highlighting the limitations of fully automated pipelines and the necessity of human verification. As shown, models may produce incorrect answers in counting such as miscounting cushions, dice pips, or food items, generate ambiguous or wrong questions where the answer is encoded in the questions themselves. These examples demonstrate that QA generation errors arise at both the question and answer levels.

\newpage

\newpage

\subsection{Prompts for Generating VQA Tasks} \label{app:data-construct}
\begin{tcolorbox}[promptbox, title={Template Prompt for Attribute Task }]
The primary goal of this task is to generate challenging compositional and multi-hop VQA questions for attribute reasoning. Analyze the given image carefully and generate five diverse attribute-based questions, each paired with a concise answer. The questions should rely on attributes such as color, size, shape, texture, behavior, or similar properties. \\

\end{tcolorbox}

\begin{tcolorbox}[promptbox, title={Template Prompt for OCR Task }]
The primary goal of this task is to generate high-quality VQA questions for OCR understanding. Analyze the image carefully and generate five diverse text-centric questions, each paired with a concise answer. Questions should require reading and reasoning over visible text in the image, such as signs, labels, logos, or printed content. They may involve exact text extraction or simple reasoning based on the text.\\
All answers must be directly grounded in clearly visible text and be unambiguous. Avoid paraphrasing for the answers.

\end{tcolorbox}

\begin{tcolorbox}[promptbox, title={Template Prompt for Identfication Task}]
The primary goal of this task is to generate high-quality VQA questions for object identification. Analyze the image carefully and generate five diverse identification questions, each paired with a concise answer.\\
Questions should require recognizing and naming objects, entities, or scene elements present in the image based on their functional role, spatial position, or contextual relationships with other objects.
\end{tcolorbox}

\begin{tcolorbox}[promptbox, title={Template Prompt for Comparison Task}]
The primary goal of this task is to generate high-quality VQA questions for comparison reasoning. Analyze the image carefully and generate five diverse comparison questions, each paired with a concise answer.\\
Questions should require comparing two or more objects, entities, or regions in the image based on attributes such as size, color, shape, quantity, position, or other visually observable properties. They may involve identifying differences, similarities, or determining which object satisfies a given condition
\end{tcolorbox}

\begin{tcolorbox}[promptbox, title={Template Prompt for Spatial Task }]
The primary goal of this task is to generate high-quality VQA questions for spatial understanding. Analyze the image carefully and generate five diverse spatial reasoning questions, each paired with a concise answer. Questions should involve spatial relations such as left/right, above/below, in front of/behind, inside/outside, near/far, between, surrounding, or relative position. 

\end{tcolorbox}

\begin{tcolorbox}[promptbox, title={Template Prompt for Counting Task }]
The primary goal of this task is to generate high-quality VQA questions for counting task. Analyze the image carefully and generate five diverse counting reasoning questions (hard and very hard), each paired with an answer. The questions may involve basic counting, attribute-aware counting, or spatial counting. Label each question with a difficulty level.\\
Difficulty Guidelines: \{\}
\end{tcolorbox}

\begin{tcolorbox}[promptbox, title={Template Prompt for Hallucination-Aware VQA }]
The primary goal of this task is to generate hallucination-based VQA questions that probe whether a model relies on non-existent or unsupported visual evidence. Analyze the image carefully and generate three hallucination questions.\\

Each question should be formulated such that it refers to objects, attributes, text, or relations that are not present in the image, or cannot be verified from the visual content, while remaining consistent with the specified task \{task\}. The goal is to test whether a model incorrectly assumes the presence of such elements.\\

Answers should be concise and reflect the absence of evidence.
\end{tcolorbox}

\subsection{Prompt for LLM-as-a-Judge Evaluation of Answer Correctness} \label{app:judge}
\begin{tcolorbox}[promptbox, title={Prompt}]
You are an evaluator for a visual question answering task. Your task is to determine whether the predicted answer is correct with respect to the ground-truth answer for the given question.\\

Evaluation rules:\\
\begin{enumerate}
    \item Accept minor wording differences, paraphrases, singular/plural variation, and equivalent expressions.
    \item Accept answers that are semantically equivalent to the ground-truth answer.
    \item Reject answers that are incomplete, overly vague, or refer to the wrong object, attribute, count number, text, or relation. 
    \item For counting questions, the numeric value must match exactly unless the reference explicitly allows a range. 
    \item For OCR questions, ignore capitalization and minor punctuation differences, but do not ignore incorrect characters or different words. 
    \item For hallucination questions, accept only answers that clearly indicate the target does not exist or is not present or the answer cannot be determined from the image.
    \item Be strict: the predicted answer should mean the same thing as the ground-truth answer in the context of the question.
\end{enumerate}

Return your decision in JSON with the following fields:\\
\{\\
  "correct": 0 or 1,\\
"reason": "short explanation"\\
\}

Question: \{question\} \\
Ground-truth answer: \{gt-answer\}\\
Predicted answer: \{pred-answer\}\\
Task type: \{task-type\}\\
Domain: \{domain\}

\end{tcolorbox}

\subsection{Structured Output Format} \label{app:output-format}

Models are required to produce structured outputs consisting of a textual answer and corresponding multi-instance grounded visual evidence in a predefined format. For models that do not natively support structured grounding outputs, predicted regions are mapped to this format via post-processing. If no grounded object is present (\eg, hallucination cases), the masks field is empty.

\begin{tcolorbox}[promptbox, title={Structured Output Template}]
\begin{lstlisting}[basicstyle=\small\ttfamily, breaklines=true]
{
   "image":
  {
    "image_id": {<id>},
    "width": H,
    "height": W,
    "file_name": "{<name>}"
  },
  "question": "<input question>",
  "prediction": {
    "text": "<final answer with optional object references>",
    "masks": [
      {"size":[H, W],"counts":"<RLE-encoded mask for instance1>"},
      {"size":[H, W],"counts":"<RLE-encoded mask for instance2>"}
      .
      .
      .
    ]
  }
}
\end{lstlisting}
\end{tcolorbox}

\end{document}